\documentclass{article}
\usepackage{acl}

\usepackage[utf8]{inputenc} 
\usepackage[T1]{fontenc}    
\usepackage{hyperref}       
\usepackage{url}            
\usepackage{nicematrix}       
\usepackage{amsfonts}       
\usepackage{nicefrac}       
\usepackage{microtype}      
\usepackage{xcolor}         
\usepackage{amsmath}
\usepackage{amsthm}
\usepackage{graphicx}
\usepackage{float}
\usepackage{caption}[justification=centering]
\usepackage{subcaption}
\usepackage{colortbl}
\usepackage{booktabs}
\usepackage{changepage}
\usepackage{listings}
\usepackage{tikz}
\usepackage{array}
\usepackage{arydshln}
\usepackage{boldline}

\usetikzlibrary{arrows.meta, positioning, automata}

\newcolumntype{B}{!{\vrule width 2pt}}

\newtheorem{theorem}{Theorem}[section]

\definecolor{dark2green}{rgb}{0.1, 0.65, 0.3}
\definecolor{dark2orange}{rgb}{0.9, 0.4, 0.}
\definecolor{dark2purple}{rgb}{0.4, 0.4, 0.8}
\newcommand{\first}[1]{\textbf{\textcolor{dark2green}{#1}}}
\newcommand{\second}[1]{\textbf{\textcolor{dark2orange}{#1}}}

\definecolor{pastel-red}{HTML}{FF9AA2}
\definecolor{pastel-green}{HTML}{B5EAD7}
\definecolor{pastel-blue}{HTML}{A7DBD8}
\definecolor{pastel-teal}{HTML}{9AD9DB}
\definecolor{pastel-yellow}{HTML}{FFFFCC}
\definecolor{pastel-purple}{HTML}{D3C0F9}
\definecolor{pastel-orange}{HTML}{FFC2A2}
\definecolor{pastel-pink}{HTML}{FFB7B2}
\definecolor{pastel-cyan}{HTML}{A1E8E2}
\definecolor{pastel-lavender}{HTML}{E0BBE4}
\definecolor{light-coral}{HTML}{F08080}
\definecolor{light-sea-green}{HTML}{20B2AA}
\definecolor{light-slate-blue}{HTML}{8470FF}
\definecolor{light-goldenrod-yellow}{HTML}{FAFAD2}
\definecolor{light-pale-green}{HTML}{98FB98}
\definecolor{light-orchid}{HTML}{E6A8D7}
\definecolor{light-salmon}{HTML}{FFA07A}
\definecolor{light-sky-blue}{HTML}{87CEFA}
\definecolor{light-steel-blue}{HTML}{B0C4DE}
\definecolor{light-turquoise}{HTML}{AFEEEE}

\definecolor{dkgreen}{rgb}{0,0.6,0}
\definecolor{gray}{rgb}{0.5,0.5,0.5}
\definecolor{mauve}{rgb}{0.58,0,0.82}

\title{Lightweight reranking for language model generations}

\author{%
  Siddhartha Jain\thanks{Corresponding author} \\
  AWS AI Labs \\
  \texttt{siddjin@amazon.com} \\
  \And
  Xiaofei Ma \\
  AWS AI Labs \\
  \texttt{xiaofeim@amazon.com} \\
  \AND
  Anoop Deoras \\
  AWS AI Labs \\
  \texttt{adeoras@amazon.com} \\
  \And
  Bing Xiang \\
  AWS AI Labs \\
  \texttt{bxiang@amazon.com} \\
}

\begin{document}

\maketitle

\begin{abstract}

Large Language Models (LLMs) can exhibit considerable variation in the quality of their sampled outputs. Reranking and selecting the best generation from the sampled set is a popular way of obtaining strong gains in generation quality. In this paper, we present a novel approach for reranking LLM generations. Unlike other techniques that might involve additional inferences or training a specialized reranker, our approach relies on easy to compute pairwise statistics between the generations that have minimal compute overhead. We show that our approach can be formalized as an extension of self-consistency and analyze its performance in that framework, theoretically as well as via simulations. We show strong improvements for selecting the best $k$ generations for code generation tasks as well as robust improvements for the best generation for the tasks of autoformalization, summarization, and translation. While our approach only assumes black-box access to LLMs, we show that additional access to token probabilities can improve performance even further. 
  
\end{abstract}

\section{Introduction}

The rapid advancement and remarkable achievements of generative large-scale pre-trained language models (LLMs) have brought about a revolutionary transformation in the field of natural language processing (NLP). These models have demonstrated significant enhancements in various NLP applications, such as machine translation, summarization, and code generation. Individual generations sampled from the models often yield high-quality results. However the quality of generated outputs can exhibit considerable variability. Multiple output samplings for the same input can produce certain generations which are of substantially higher quality than the quality of the average generation from the model.

Several approaches have been proposed to exploit this phenomenon. One strategy involves improving the underlying models themselves to make the quality of the average generation consistently better. This can be achieved by taking existing model generations, ranking them based on a human feedback, automated evaluation metrics like BLEU score, or execution feedback in case of code. The ranked generations can then be finetuned on directly or can be used to train a reward model that can be used in an RL loop~\citep{hsieh2023distilling, ouyang2022training, ho2022large, polu2022formal, liu2021simcls, ouyang2022training}. Another common approach is best-of-$n$ sampling or \textit{reranking}. In this approach, the underlying model is not touched -- we instead take multiple samples from the model and select the best one post-facto using a reranking method~\citep{ravaut2022summareranker, jiang2022pairreranker, zhang2022coder, chen2021evaluating, shi2022natural, li2022competition, mizumoto2016discriminative, uesato2022solving}. While this approach can often given strong improvements, most extant reranking techniques involve computationally intensive or cumbersome methods to compute the ranking criterion. These include methods like training an auxiliary model as a reranker, evaluating the probability of the query given the generated answer (query likelihood) but at the price of doubling the inference cost, etc. In case of code generation models, another alternative is executing the generated code on unit tests. While such an approach has been applied in various models such as AlphaCode~\citep{li2022competition} which is targeted towards contest coding problems, it becomes much less feasible as you move past the contest coding setting due to the complexity of setting up the  build environment for arbitrary code as well as sandboxing it appropriately.

Recently, a simple approach, called self-consistency was proposed for selecting the best answer from multiple generations~\citep{wang2022self} for tasks where the set of possible answers is small -- for example multiple choice questions or math word problems 
where there is a \textit{unique} answer consisting of a single or a very limited number of tokens. In that paper, the authors sample multiple chain-of-thought generations from the LLM, extract the predicted answer at end each generation and select the answer with the most number of votes. The motivation behind this is the observation that you can take \textit{different} reasoning paths to get to the same answer. Thus the method aims to \textit{marginalize} over multiple different reasoning paths and rank the answers based on their \textit{marginal} probability rather than their probability conditioned on a single reasoning path. While they achieve substantial improvements over existing baselines, it is not immediately clear how to apply this to open-ended generation tasks like code generation, summarization, or translation - where there is often no chain-of-thought or reasoning path to marginalize over, nor is there necessarily a unique correct answer. 

We start off with two key observations -- (1) We can have semantically equivalent or near-equivalent generations that are nevertheless not exact matches. These are one subset of generations we can marginalize over (2) For open-ended tasks, a generation can encompass multiple elements. For summarization, there might be multiple relevant facts in the text that a good summary should mention. For code, there might be multiple branch conditions that need to be present to generate a correct implementation. Our generation set could be structured such that while different generations include a different subset of elements (different facts in case of summarization or different branch conditions in case of code), we have only a single generation that contains \textit{all} of the relevant elements. In this case, simply marginalizing over semantically equivalent generations would not be sufficient as there is no semantically equivalent generation for the optimal generation. 

We develop these two observations in the next section into a minimal overhead reranking method for such open-ended tasks which does not require access to token probabilities.

Concretely, our contributions are as follows --

\begin{itemize}
    \item We connect the above two observations with the notion of self-consistency. Based on that connection, we then proceed to design an effective minimal overhead reranker which does not require access to token probabilities. We show that the reranking methods utilized in previous works~\citet{shi2022natural, li2022competition} can also be understood within the same conceptual framework.
    \item We conduct simulations where we demonstrate that our framework is capable of recovering the best or near-best generation in many cases. We also prove some properties of our methodology that provide guarantees on its effectiveness.
    \item We extend our reranker to \textit{optionally} account for token log probabilities (if they are provided) and show that doing so gives a much better reranker than just mean log probability reranking (which also requires access to token log probabilities)
    \item Empirically, while our focus is on code generation tasks where we demonstrate significant gains, we also experiment with the tasks of autoformalization, summarization, and translation and find that our approach leads to non-trivial though smaller gains there.
    \item As our method is based on pairwise similarity between generations, we are able to leverage that property to improve ranked best-of-$k$ performance for different values of $k$.
    \item We conduct multiple experiments ablations to understand the effect of various experimental settings.
\end{itemize}

The rest of the paper is organized as follows. In Section~\ref{motivation} we present our motivation. In Section~\ref{method} we present our method and the similarity function. In Section~\ref{results}, we present and discuss our experimental results. In Section~\ref{background}, we describe the related work and we finally conclude in Section~\ref{conclusion}.

\section{Motivation \label{motivation}}

Consider the following coding problem from the MBPP dataset -- 

\begin{minipage}{\linewidth}
\begin{lstlisting}
def remove_dirty_chars(string, second_string):
    """
    Write a function to remove characters from the first string which are present in the second string.
    >>> remove_dirty_chars("probasscurve", "pros")
    'bacuve'
    >>> remove_dirty_chars("digitalindia", "talent")
    'digiidi'
    >>> remove_dirty_chars("exoticmiles", "toxic")
    'emles'
    """
\end{lstlisting}
\end{minipage}

A solution to the above problem would be semantically equivalent to "iterate over the string skipping characters in second\_string and then convert the result back to a string and return". Two parts of the semantic meaning of this solution could then be (1) the return type should be a string (2) when iterating through the string, any character in second string has to be skipped over. These observations can be converted into \textit{predicates} for the generations. Specifically, for this prompt, we can define the predicates (1) $p_1=$ is the return value of the generated program a string? (2) $p_2=$ in the generated program, are all characters in \texttt{second\_string} skipped over in the return string? These predicates capture properties of the semantic meaning of the generations. We sample three generations for the above prompt resulting in the following generations:

\begin{minipage}{\linewidth}
\begin{lstlisting}
# First generation (Incorrect)
    return [char for char in string if char not in second_string]

# Second generation (Incorrect)
    return ''.join([char for char in string])

# Third generation (Correct)
    return ''.join([char for char in string if char not in second_string])
\end{lstlisting}
\end{minipage}

Now if we were able to evaluate the above predicates at inference time on the generations, we would be able to detect that generation 3 is the only one that satisfies both and is thus an optimal generation. However generating the relevant predicates, and then generating code to evaluate arbitrary predicates on code that confirms to the given natural language specification with high precision is an unsolved problem.

Is there a way transform the problem into something more tractable? Let us look at the votes each predicate gets from the different generations (i.e. on how many generations the predicate evaluates to true). $p_1$ gets 2/3 votes (from the 2nd and 3rd generation) and thus the \textit{majority vote} is that it should be true. $p_2$ gets 2/3 votes (from the 1st and 3rd generation) and thus the majority vote again says it should be true. Generation 3 is the only one that agrees with the majority vote for $p_1, p_2$ and is thus the consensus choice. 

\begin{figure*}[ht]
\captionsetup{justification=centering}
  \centering
  \includegraphics[width=0.7\textwidth]{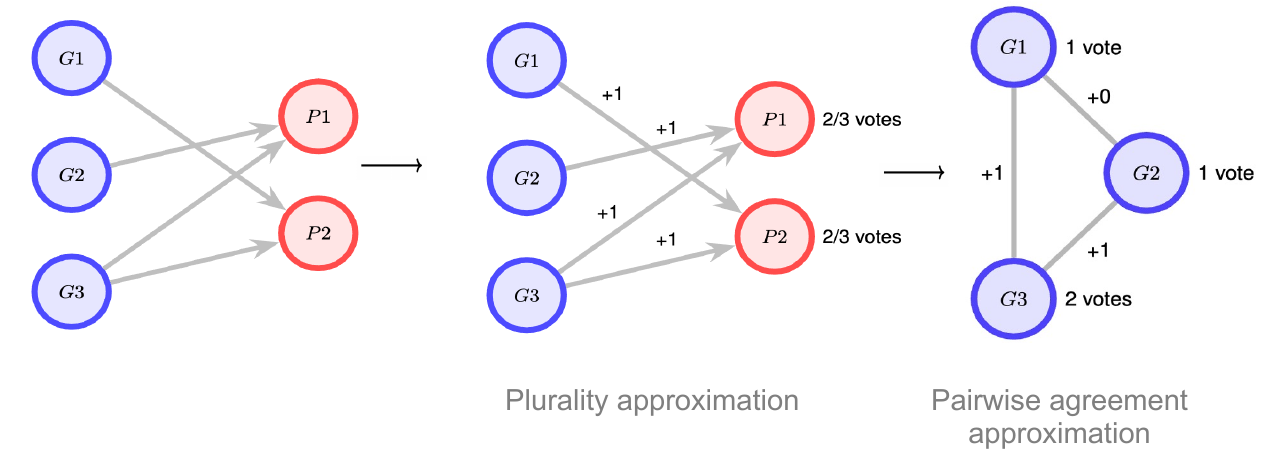}
  \caption{\scriptsize On the left we have the original setup where we have predicates which we know the optimal generation should satisfy and which we can evaluate on the generations. In the middle, we drop the assumption that we know whether the optimal generation should satisfy the predicates or not. On the right, we drop the assumption that we need to evaluate the predicates on the different generations -- only assuming we know on how many predicates a pair of generations agree}
  \label{fig:transformation}
\end{figure*}

In fact, we do not even have to do the step of first counting votes for $p_1, p_2$ to figure out what their majority vote value is! We can just compute how much a generation agrees with the other 2 generations on the evaluation for $p_1, p_2$. To elaborate, generation 1 agrees with generation 3 on $p_2$ but not $p_1$. It does not agree with generation 2 on anything. Thus it has a total agreement score of 1. Similarly generation 2 also has an agreement score of 1. Generation 3 however agrees with generation 1 on $p_2$ and with generation 2 on $p_1$ resulting in an agreement score of 2. Thus generation 3 has the highest agreement with all other generations and is the consensus choice. This transformation is depicted in Figure~\ref{fig:transformation}.

There are a couple of points to note regarding the above. (A) The answer we want for both predicates ends up being also what the majority vote predicts. This is the connection to self-consistency that we will formalize shortly. (B) While the above procedure handles Observation (2) in the mentioned in the introduction, does it also handle Observation (1)? Note that if we had a 4th generation 

\begin{minipage}{\linewidth}
\begin{lstlisting}
# Fourth generation (Correct)
    new_str = ''
    for c in string:
        if c not in second_string
            new_str += c
    return new_str
\end{lstlisting}
\end{minipage}

this would also agree with the 3rd generation on $p_1, p_2$. Thus this generation is semantically equivalent to the 3rd generation with respect to $p_1, p_2$. Including this generation would lead to a tie between the 3rd and 4th generation with respect to the scheme above. Thus the above procedure can also account for generations that are semantically equivalent.

As mentioned before, at inference time however, we have access neither to such predicates nor to what their evaluation on the different generations might be. However, as the previous example showed, we do not actually even need an evaluation of the predicates on the generations. We only actually need an understanding of how much a pair of generations agree on relevant predicates. In the next section we will describe simple similarity functions to approximate this agreement without any knowledge of the possible predicates and later show their empirical effectiveness. For now, we attempt to formalize the above intuition, connect it with self-consistency, and give it theoretical grounding.

Let $\mathbf{v}$ be a vector of length $k$ where each element represents a predicate and the value represents the property's value that you want the generation to have. For example, for the example above, $\mathbf{v}$ would be of length 2 with a value of 1 in both. In general, we do not restrict $v$ to be just a binary vector as some predicates can have multiple possibilities as an answer. Let $\mathbf{u}_i$ be the vector for the $i$th generation representing the predicates' values for that generation. We now make the \textit{self-consistency} assumption which is that for each individual predicate, the most frequent response is assumed to be correct. Formally if $\mathbf{v}^l$ can take on $m_l$ values ${1,\dots,m_l}$ and without loss of generality, $\mathbf{v}^l = 1$, then $l = \arg \max_j \sum_{i=1}^n \mathbb{I}(u^l_i = j)$.

Now as mentioned, at inference time, we may not have access to either the predicates or their values for different generations. Thus we only assume access to the \textit{agreement} between the different generations on the predicates' values (later on we will approximate this with similarity functions between generations). In particular we assume we know what the pairwise fractional agreement on the predicates is between generations denoted as $a(\mathbf{u}_i, \mathbf{u_j}) = \frac{1}{k} \sum_{t=1}^k \mathbb{I}(\mathbf{u}^t_i = \mathbf{u}^t_j) \forall i, j \in [1, n]$ where $i$ indexes the generations and $t$ the predicates. We then try to identify a generation $i$ such that the average pairwise fractional agreement for that generation with all other generations is maximized -- i.e. $a(\mathbf{u}_i, \mathbf{v})$ is maximized. 

Given this problem formulation and selection criterion, we can establish the following:

\begin{theorem}
For $k = 1$, we always recover the best $\mathbf{u}$. However for $k > 1$, it is not guaranteed.
\end{theorem}

Informally if there is only a single predicate we care about, then we can always recover the optimal generation. Moreover:

\begin{theorem}
If there exists $\mathbf{u}_b = v$, then $b = \arg \max_i \frac{1}{n-1} \sum_{i \neq j} a(\mathbf{u}_i, \mathbf{u_j})$.
\end{theorem}

Informally this says that if a generation $g$ exists such that its predicate vector perfectly aligns with the optimal vector $v$, selecting the generation with the highest average fractional agreement with other generations will pick $g$.

The previous theorem only works if the optimal generation is part of the set of generations. What if that is not the case (as is likely)? The next theorem gives upper and lower bounds on the fractional agreement we can expect. Now if we assume that $\mathbf{u}^j_i$ are iid from $Bernoulli(p_j)$ and $n$ is the number of generations, then we can show that

\begin{theorem}
$\mathbb{E}[\sum_j^k \mathbf{u}^j_b] \leq \sum_{j=1}^k p_i + \sqrt{\frac{k\log{n}}{2}}$
\end{theorem}

where $\mathbf{u}_b$ denotes the sequence selected by our method.

All proofs for these theorems are presented in the Supplement. While the theorems give some guarantees on the performance, the bounds in Theorem 2.3 are still not very tight. Furthermore, They are only for the case where the predicates are binary valued. To further substantiate our selection criterion — picking the generation with the highest average fractional agreement with all other generations — we conducted a simulation. The setup is as follows -- we fix the number of predicates (length $k$ of the vector $\mathbf{v}$ in the above notation) as well as the number of values the predicate can take. We then simulate the generations predicate evalutions by assuming a generation has an equal chance of having an value the predicate can take. However we force the self-consistency constraint that for every predicate, the \textit{plurality} of generations should have the property that matches the predicate value in $v$. The results are in the Supplement. Our findings show that our method successfully recovers the best generation the majority of the time, significantly outperforming random selection. Moreover, on average, the generation we recover demonstrates nearly 100\% agreement with best generation, even in cases where we do not select the best generation. The full details are in the Supplement.

\section{Method \label{method}}
As previously mentioned, we may not have the capability to compute predicates at inference time, thereby rendering the computation of the exact fractional agreement with $\mathbf{v}$ i.e. $a(\mathbf{u}, \mathbf{v})$, unattainable. However as we found out in the last section, choosing the generation that has the maximum average fractional similarity with all other generations can be a good approximation. However as we may not have predicates at inference time, we cannot always compute that either. Intuitively however, if two generations are more similar to each other -- for an appropriate definition of similarity -- then they will tend to agree more on \textit{any} possible predicates. Surprisingly, we find that a very simple similarity function, which we will define shortly, is sufficient for our purposes.

Once we have our similarity function, we can define a generalized self-consistency score $GSC_{Sim}(i)$ for each generation $i$, given by $\frac{1}{M-1} \sum_{j=1, j\neq i}^M Sim(i, j)$. Here, $Sim$ denotes the similarity function, and $M$ represents the number of generations.

For generations with unique answers, if we have: \\

$Sim(i, j) = \mathbb{I}($Answer in generation $i$ is an exact match with Answer in generation $j)$ \\

this is equivalent to the self-consistency criterion. Two other reranking methods - MBR-Exec~\citep{shi2022natural} and AlphaCode~\citep{li2022competition} - can be viewed in terms of the same formulation with the difference being that of the similarity function. MBR-Exec \textit{executes} model generated code. It then defines gives a similarity score of $1$ if a pair of programs agree on \textit{all} unit tests and $0$ otherwise
For each program, they sum the similarity vs all other programs and pick the program with the highest similarity. Similarly AlphaCode clusters its generated programs by executing them on test cases and selecting a program from the largest cluster -- with two programs cluster together if they agree on on all test cases. This is conceptually equivalent to what MBR-Exec does. We give further evidence that this is a useful way to frame self-consistency by evaluating another OpenAI Ada embedding based similarity function (Section~\ref{ada_embed_analysis} in the Supplement). While its performance is promising, as the similarity function is a lot more heavyweight requiring a separate embedding model, we chose not to explore it further.

\begin{figure*}[ht]
\captionsetup{justification=centering}
  \centering
  \includegraphics[width=0.6\textwidth]{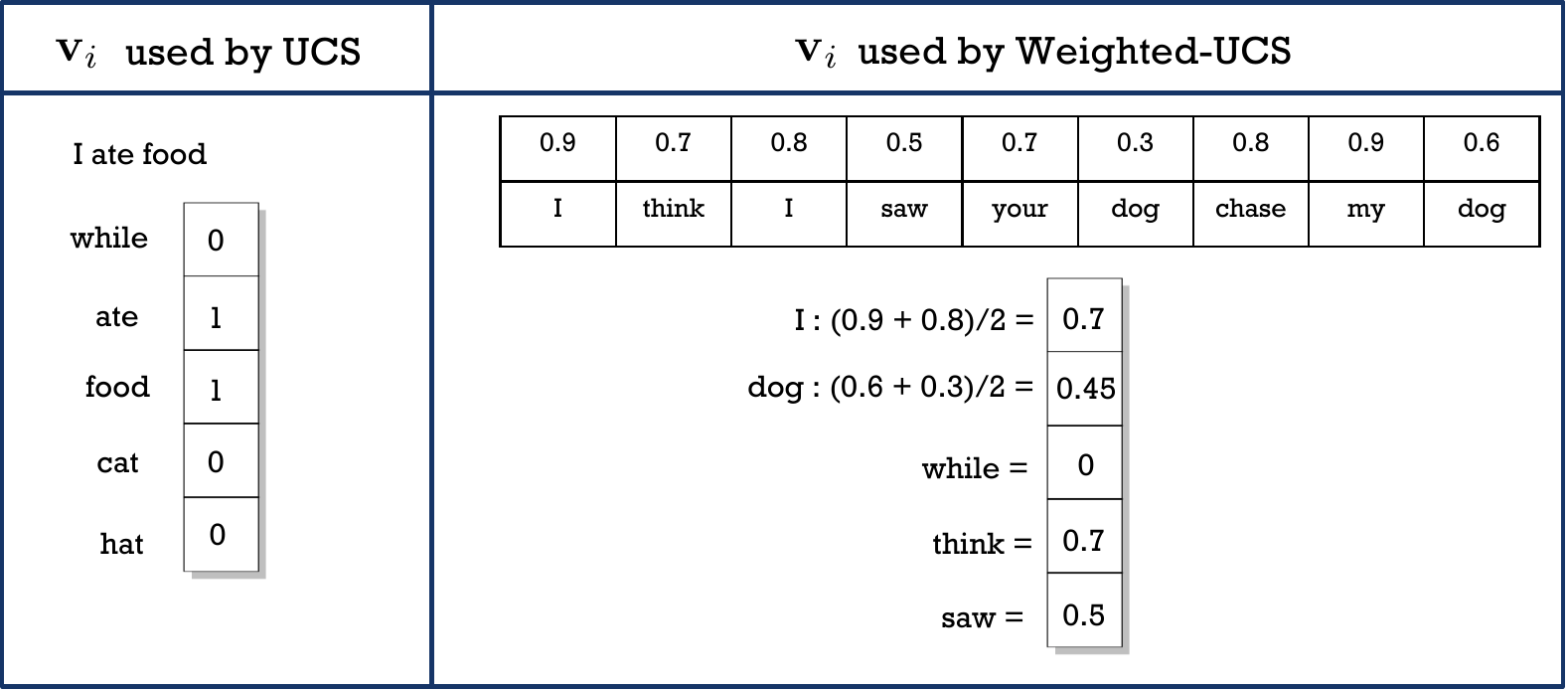}
  \caption{\scriptsize On the left, we depict the $\mathbf{v}_i$ corresponding to the sentence \texttt{I ate food} used by UCS. On the right we show the $\mathbf{v}_i$ corresponding to the sentence \texttt{I think I saw your dog chase my dog} used by Weighted-UCS}
  \label{fig:ucs_weighted_ucs}
\end{figure*}

One straightforward way to encode a generation is by using a binary vector that denotes the presence or absence of an n-gram. Surprisingly, we find this simple encoding to be sufficient for defining a robust similarity function. For open-ended generation, we define our similarity function as follows. For each generation we define a vector $\mathbf{v}$ of size $|V|$ where $V$ is set of all possible n-grams for $n=1$ to $n=K$ where $K$ is a hyperparameter. For the experiments in this paper, we simply use $K=1$. We show in Section~\ref{ngram_varying}, increasing $K$ can be helpful though only up to a point. Each element $i$ of $\mathbf{v}$ is simply whether token $i$ is present in the generation or not. We then take the inner product between two such vectors as similarity. We call this the Ngram consistency score (NCS) and refer to the $K=1$ version as the Unigram consistency score (UCS). Figure~\ref{fig:ucs_weighted_ucs} shows a visualization of $\mathbf{v}$ for an example sentence. Formally

\begin{equation*}
    UCS(i, j) = \frac{1}{|V|} \mathbf{v}_i\cdot \mathbf{v}_j
\end{equation*} where

\begin{equation*}
    \mathbf{v}^j_i = \mathbb{I}(t_j \in g_i)
\end{equation*}

where $t_j$ is the $j$th token and $g_i$ the $i$th generation. This definition only requires model generations and incurs minimal computational overhead -- we only need to compute the unigram overlap instead of training an auxiliary model, running generated programs, or performing additional inferences using the same model (which will increase compute cost as well as latency). Notably, we don't normalize the inner product by the norm of the vectors. This is a deliberate design choice that encourages more diverse sequences, in response to known issues of neural generation models producing degenerate and repetitive sequences~\cite{zhang2022coder, welleck2019neural}. We delve into this topic in Section~\ref{normalization_analysis} in the Supplement. 

When token probabilities are available, we can leverage them to improve our approach. Intuitively, if a generation has a low token probability for the generated token, then finding a match for that that token should count for less. In accordance with this intuition, we introduce two further variants. First we modify the  definition of $\mathbf{v}$ as follows

\[
    \mathbf{v}^j_i = 
    \begin{cases} 
    \frac{1}{c_j^i} \sum_{k}^{c_j^i} p(t_j^{i,k}) & \text{if $t_j \in g_i$,} \\
    0 & \text{otherwise}
    \end{cases}
\]

where $c^j_i$ is the number of times token $t_j$ appears in generation $i$ and $p(t_j^{i,k})$ is the token probability of the $j$th token's $k$th appearance in generation $i$. We call this the weighted n-gram consistency score (WUCS). Figure~\ref{fig:ucs_weighted_ucs} has a visualization for an example sentence.

The mean log probability of a sequence is an oft-used ranking method. We can combine it with WUCS by further weighting each generation by the per token probability as follows -- for a generation $i$, $\text{Consensus-WUCS} = WUCS\cdot e^{(1/|g_i|)\cdot p(g_i)}$ where $g_i$ is the length of generation $i$. 

Finally, to rank the generations, we employ $\arg \max_i GSC_{Sim}(i)$ where $Sim$ can take the form of UCS, WUCS, or Consensus-UCS.

\subsection{Extending to ranked $pass@k$}

A common evaluation metric for code generation problems is ranked $pass@k$ wherein we assess whether \textit{any} program among the top $k$ selected programs (selected from a larger set) can pass all the given unit tests for that problem. Typically, the top $k$ generations are selected based on a predetermined ranking. However, with our similarity-based metric, we can apply a more nuanced approach. 

For a particular problem, if the highest-ranked generation for a specific prompt is correct, we have already succeeded. We would only need to utilize the remaining generations in our $k$-budget if the top-ranked generation does not pass some unit test case. In this event, we could consider the top-ranked generation as a hard negative and select the next generation that exhibits lower similarity to the top-ranked generation.

More specifically, if we have selected programs $S_{k'}$ so far ($|S_{k'}| = k' < k$, then we modify the GCS function to select the $k'+1$th item in the list. In particular, we compute

\[
    GCS^{ranked}_{sim} = \frac{1}{n-1}(\sum_{j \notin S_{k'}} sim(i, j) - \sum_{j \in S_{k'}} sim(i, j))
\]

Note that for $k=1$, $GCS$ and $GCS^{ranked}$ are equivalent. We demonstrate in Section~\ref{pass_at_k_compare} that $GCS^{ranked}_{Sim}$ performs significantly better in ranking for $pass@k$ where $k > 1$ than raw $GCS$. This approach leads to a more efficient utilization of the ranked generations, improving the overall effectiveness of the code generation task.

\section{Results\label{results}}

We conducted experiments utilizing the Codex family of models, specifically Codex-davinci-001, Codex-davinci-002, and Codex-Cushman as well as Llama family of models. In addition we also evaluated GPT-J for Xsum, MiniF2F, and WMT14. Unfortunately due to the unexpected shutdown of the OpenAI API, we were unable to obtain results for Codex-001 and Codex-Cushman on the Xsum, MiniF2F, and WMT14 datasets. We evaluated these models on a range of datasets for code generation tasks -- in particular on the HumanEval~\citep{chen2021evaluating}, MBPP, MBPP-sanitized~\citep{austin2021program} datasets for code generation. For the autoformalization of MiniF2F to Isabelle, we used the dataset provided by~\citep{jiang2022draft}. For text summarization, we utilized the Xsum dataset~\citep{narayan2018don}. For machine translation, we used the WMT14 French to English and German to English datasets~\citep{bojar2014findings}.

Our primary evaluation metric for code generation is ranked $pass@1$ where we rerank a sample set of generations and assess whether the top-ranked generation successfully passes all unit tests. We also evaluate with ranked $pass@k$ for $k>1$. For the MiniF2F autoformalization task, we measure the quality using the BLEU score, following~\citet{wu2022autoformalization}. For Xsum we use the Rouge-2 and Rouge-L scores for evaluation. For all code generation datasets, we sample 125 generations from the models which serves as our dataset for the different experiments
For MiniF2F and Xsum, we sample 50 generations from the model. Unless otherwise specified, for all experiments, we use the Codex-davinci-002 model. Following~\cite{shi2022natural, zhang2022coder}, we perform bootstrap sampling 50 times with a sample size of 25 to generate the results. 

Our baselines are Random selection, Ranking by mean log probability, Ranking using Medoid in our confidence weighted unigram space, and for code generation - ranking using the Coder Reviewer Ranker method~\citep{zhang2022coder}. \textbf{A full description of the datasets, experiments, and the baselines is in the Supplement. We also have several additional results in the Supplement.}

\subsection{GSC scores are higher for correct answers}

As a sanity check, we first evaluate whether the GSC scores are indeed higher for the correct generations~\footnote{We used the generations in~\cite{li2022competition} provided by them as part of their Supplementary Material.} The results are in Table~\ref{table:UCS_correct_incorrect} in the Supplement. The ratios are consistently $>1$ for all models except for the UL2-20B model for which they still remain very close to 1. 

\subsection{UCS shows strong improvements for Code Generation}

As shown in Tables~\ref{table:accuracy} and~\ref{table:code_gen_mrr} (Supplement), the application of the UCS, WUCS, and Consensus-WUCS methods leads to substantial improvements in the accuracy as well as mean reciprocal rank of code generation across various models and datasets. 

In the HumanEval dataset, UCS variants consistently outperform the traditional methods, namely Random and mean log probability. For instance, the Codex002 model exhibits a substantial accuracy improvement from 0.435 (Random) to 0.568 (Consensus-WUCS). Even the less performing models, such as Llama-13B and Llama-30B, exhibit noticeable accuracy gains when our proposed methods are employed.

Similar trends are observed in the MBPP-S and MBPP datasets. UCS, WUCS, and Consensus-WUCS consistently improve the accuracy across all models. Specifically, the Consensus-WUCS method consistently dominates Random and mean log probability ranking in all categories, and almost always outperforms WUCS as well. Of particular note is the performance of WUCS, which surpasses the mean log probability method in every model and dataset combination. In fact it is the best method for all dataset and model combinations except LLama-13B model for MBBP and MBPP-S. UCS, which does not require token probabilities and relies only on the generations, also demonstrates a consistent superiority over the random reranking.

Consensus-WUCS and WUCS are also almost always better than the Medoid based approach with Consensus-WUCS outperforming it 13/15 times. A discussion of the mean reciprocal ranking performance is deferred to the Supplement but the trend is similar.

\begin{table*}[htb]
\hspace{0.2in}\begin{minipage}{0.5\textwidth}
\centering
\fontsize{8.25pt}{8.25pt}\selectfont
\setlength\tabcolsep{6.95pt} 
\scalebox{0.9}{
\begin{tabular}{p{4cm}ccccc}
\toprule
{\scriptsize \textbf{No logprobs used}} \\
\toprule
 & \textbf{Random} & \textbf{UCS} \\
\midrule
 & \textbf{HumanEval}\\
\textbf{Codex002} & 0.435 & \textit{0.539} \\
\textbf{Codex001} & 0.345 & \textit{0.402} \\
\textbf{Code-Cushman} & 0.311 & \textit{0.353} \\
\textbf{Llama-13B} & 0.142 & \textit{0.177} \\
\textbf{Llama-30B} & 0.207 & \textit{0.257} \\
\midrule
 & \textbf{MBPP-S}\\
\textbf{Codex002} & 0.55 & \textit{0.572} \\
\textbf{Codex001} & 0.494 & \textit{0.523} \\
\textbf{Code-Cushman} & 0.436 & \textit{0.457} \\
\textbf{Llama-13B} & 0.247 & \textit{0.261} \\
\textbf{Llama-30B} & \textit{0.325} & 0.253 \\
\midrule
 & \textbf{MBPP}\\
\textbf{Codex002} & 0.536 & \textit{0.58} \\
\textbf{Codex001} & 0.475 & \textit{0.505} \\
\textbf{Code-Cushman} & 0.305 & \textit{0.386} \\
\textbf{Llama-13B} & \textit{0.185} & 0.183 \\
\textbf{Llama-30B} & 0.262 & \textit{0.276} \\
\\ 
\end{tabular}
}
\end{minipage}\hspace{-0.2in}%
\begin{minipage}{0.5\textwidth}
\centering
\fontsize{8.25pt}{8.25pt}\selectfont
\setlength\tabcolsep{6.25pt} 
\scalebox{0.9}{
\begin{tabular}{cccccc}
\toprule
{\scriptsize \textbf{logprobs used}} \\
\toprule
\textbf{Medoid} & \textbf{Mean-logp} & \textbf{WUCS} & \textbf{Consensus-WUCS}\\
\midrule
 & \\
0.437 & 0.539 & \second{0.558} & \first{\textit{0.568}}\\
0.354 & 0.408 & \second{0.426} & \first{\textit{0.445}}\\
0.335 & 0.355 & \second{0.373} & \first{\textit{0.381}}\\
0.17 & 0.17 & \second{0.187} & \first{\textit{0.192}}\\
0.225 & 0.228 & \second{0.263} & \first{\textit{0.267}}\\
\midrule
 & \\
\second{0.583} & 0.57 & 0.580 & \first{\textit{0.589}}\\
0.532 & 0.515 & \second{0.535} & \first{\textit{0.546}}\\
0.467 & 0.456 & \second{0.472} & \first{\textit{0.488}}\\
\first{0.284} & 0.27 & 0.266 & \second{\textit{0.277}}\\
0.357 & 0.348 & \second{0.363} & \first{\textit{0.373}}\\
\midrule
 & \\
0.563 & 0.512 & \second{0.587} & \first{\textit{0.594}}\\
0.505 & 0.503 & \second{0.520} & \first{\textit{0.525}}\\
0.343 & 0.319 & \second{0.405} & \first{\textit{0.420}}\\
\first{\textit{0.202}} & 0.197 & 0.195 & \second{0.199}\\
0.276 & 0.273 & \second{0.287} & \first{\textit{0.294}}\\
\\ 

\end{tabular}
}
\end{minipage}
\begin{minipage}{0.5\textwidth}
\centering
\fontsize{8.25pt}{8.25pt}\selectfont
\setlength\tabcolsep{6.25pt} 
\scalebox{0.9}{
\begin{tabular}{p{4cm}cc}
\toprule
{\scriptsize \textbf{No logprobs used}} \\
\toprule
 & \textbf{Random} & \textbf{UCS} \\
\midrule
 & \textbf{MiniF2F}\\
\textbf{Codex002} & 55.8 & 55.6 \\
\textbf{Llama-13B} & 24.3 & 24.6 \\
\textbf{Llama-30B} & \second{26} & 25.6 \\
\textbf{GPT-J} & 24.2 & 24.7 \\
\midrule
 & \textbf{Xsum Rouge2}\\
\textbf{Codex002} & 19.7 & 21 \\
\textbf{Llama-13B} & 9.2 & 10.4 \\
\textbf{Llama-30B} & 10.7 & 12.1 \\
\textbf{GPT-J} & 6.5 & \first{7.1} \\
\midrule
 & \textbf{Xsum RougeL} \\
\textbf{Codex002} & 33.9 & 34.8 \\
\textbf{Llama-13B} & 19.6 & 20.9 \\
\textbf{Llama-30B} & 21.4 & 23 \\
\textbf{GPT-J} & 17.2 & \first{18} \\
\midrule
& \textbf{WMT14 French $\rightarrow$} \\ & \textbf{English BLEU} \\
\textbf{Codex002} & 34.7 & 36.2 \\
\textbf{Llama-13B} & 4.3 & 4.4 \\
\textbf{Llama-30B} & 3.9 & \second{4} \\
\textbf{GPT-J} & 3.8 & \second{3.9} \\
\midrule
& \textbf{WMT14 German $\rightarrow$} \\ & \textbf{English BLEU} \\
\textbf{Codex002} & 30.7 & 31.6 \\
\textbf{Llama-13B} & 3.4 & 3.5 \\
\textbf{Llama-30B} & 3.7 & 3.7 \\
\textbf{GPT-J} & 3.1 & \first{3.3} \\
\bottomrule
\end{tabular}
}
\end{minipage}
\begin{minipage}{0.5\textwidth}
\centering
\fontsize{8.25pt}{8.25pt}\selectfont
\setlength\tabcolsep{6.25pt} 
\scalebox{0.9}{
\begin{tabular}{ccccc}
\toprule
{\scriptsize \textbf{logprobs used}} \\
\toprule
 \textbf{Medoid} & \textbf{Mean-logp} & \textbf{WUCS} & \textbf{Consensus-WUCS}\\
\midrule
 & \\
\first{58.2} & 52.9 & 55.8 & \second{56.2} \\
\first{24.9} & 24.2 & 24.7 & \second{24.8}\\
\first{26.4} & 25.6 & 25.7 & 25.7\\
\first{24.8} & 24 & \first{24.8} & \first{24.8}\\
\midrule
 & \\
\second{21.8} & 21.4 & 21.5 & \first{21.9}\\
10.3 & 10.3 & \first{10.6} & \first{10.6}\\
12 & 12.2 & \second{12.2} & \first{12.3}\\
6.9 & 6.6 & \second{7} & 6.9\\
\midrule
 & \\
\first{36.3} & 35.1 & 35.3 & \second{35.6}\\
20.7 & 20.3 & \first{21} & \second{20.9}\\
22.7 & 22.8 & \first{23.1} & \first{23.1}\\
17.5 & 16.6 & \second{17.8} & 17.5 \\
\midrule
& \\
& \\
35.9 & \second{36.6} & 36.5 & \first{37} \\
4.2 & \second{4.5} & \second{4.5} & \first{4.6} \\
\second{4} & \second{4} & \first{4.1} & \first{4.1} \\
3.8 & \second{3.9} & \first{4} & \first{4} \\
\midrule
& \\
& \\
31.2 & \second{33.2} & 32.1 & \first{34} \\
3.1 & \first{4} & 3.5 & \second{3.6} \\
3.5 & \first{3.9} & \second{3.8} & \first{3.9} \\
\second{3.2} & \second{3.2} & \first{3.3} & \first{3.3} \\
\bottomrule
\end{tabular}
}
\end{minipage}
\captionsetup{justification=centering}
\caption{\small{Accuracy of generated code for HumanEval, MBPP, MBBP-S} as well as performance on Xsum, MiniF2F, WMT14 French to English, and WMT14 German to English datasets. All scores are out of 100. Best results are colored in \first{first}, \second{second}. Italics for best in category (logprobs used vs not).}\label{table:accuracy}
\end{table*}

\subsection{UCS shows consistent improvements for non-coding tasks}

Next, we evaluated the performance of UCS, WUCS, and Consensus-WUCS on the non-coding tasks. In the case of the MiniF2F dataset, evaluated using the BLEU metric, Consensus-WUCS outperforms all other methods for the Codex002 model except for Medoid. For the Llama-13B, Llama-30B, and GPT-J models, the top performers are closely matched, with Consensus-WUCS, WUCS, and UCS all delivering competitive scores.

Turning to the Xsum dataset, we see a similar trend. For the Rouge-2 metric, Consensus-WUCS achieves the highest score for the Codex002 and both LLama models, and ties for the best score with WUCS for the Llama-13B model. In the GPT-J model, UCS performs slightly better than the WUCS and Consensus-WUCS. Nonetheless, all these methods surpass Random, and Mean-logp reranking methods and almost always surpass Medoid.

With the Rouge-L metric, UCS variants show the best performance for the all models except Codex002. For the Llama-30B model, WUCS and Consensus-WUCS share the top spot, while UCS achieves the best score for the GPT-J model. Once again, these methods generally outperform Medoid, Random, and Mean-logp reranking methods. 

For the WMT14 translation dataset, Conensus-WUCS is the best for all models and both tasks except for the German to English Llama-13B model. WUCS also shows strong performance, matching Consensus-WUCS score on 3 model-task combinations. UCS is also consistently better than random selection. 

In total, Consensus-WUCS gets the top spot in 12/20 comparisons, WUCS in 7/20, UCS in 3/20, and Medoid in 5/20 primarily due to MiniF2F.

\subsubsection{Cause of smaller performance improvements for non-coding tasks}

We took the top 3 and bottom 3 generations for coding and non-coding tasks and computed the unigram overlap within each set. The results are in Table~\ref{table:diversity} (Supplement). The ratio of the overlap for coding tasks was a lot higher than that for non-coding tasks giving a hint as to why we see a much stronger improvement for coding tasks. This means that if a unigram is not shared between two generations, that gives a lot more information about whether two generations are semantically far apart for coding tasks versus non-coding tasks. Thus computing the unigram overlap statistic is a lot more informative for code generation tasks vs non-code generation tasks which could be the reason behind the smaller gains for non-coding tasks. However we want to note that while the gains are smaller, they are similar to gains that that past published papers report for such metrics and importantly, the gains are robust across different tasks and models. 


\subsection{$GCS^{ranked}$ comparison\label{pass_at_k_compare}}

\begin{figure}[ht]
\captionsetup{justification=centering}
  \centering
  \begin{minipage}[b]{0.33\textwidth}
  \includegraphics[width=\textwidth]{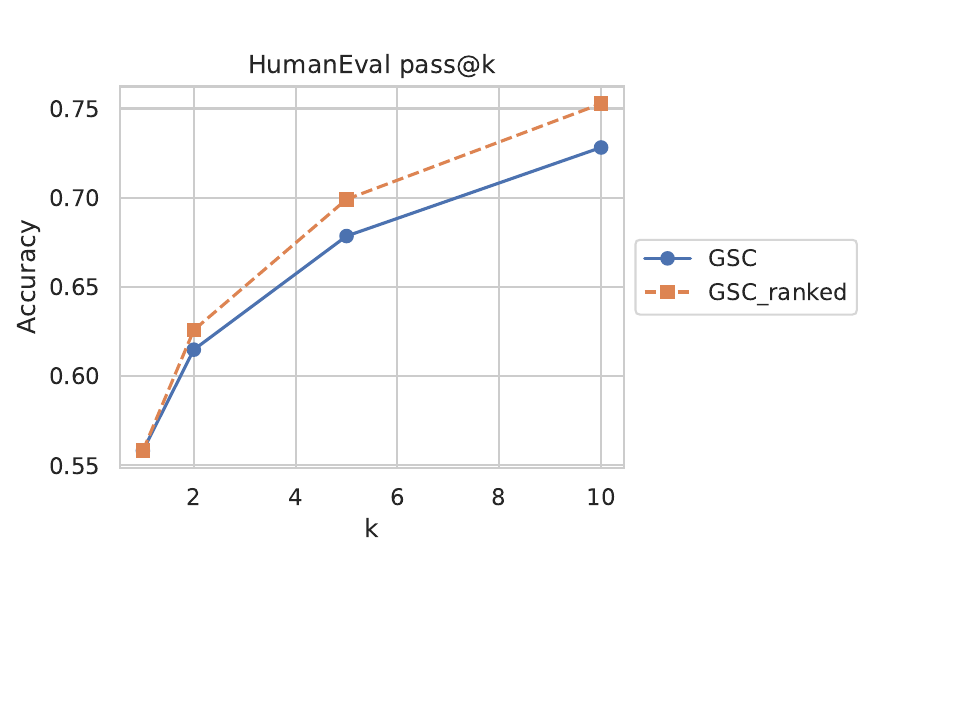}
  \end{minipage}
  \hspace{-0.35in}
  \hfill
  \begin{minipage}[b]{0.33\textwidth}
  \includegraphics[width=\textwidth]{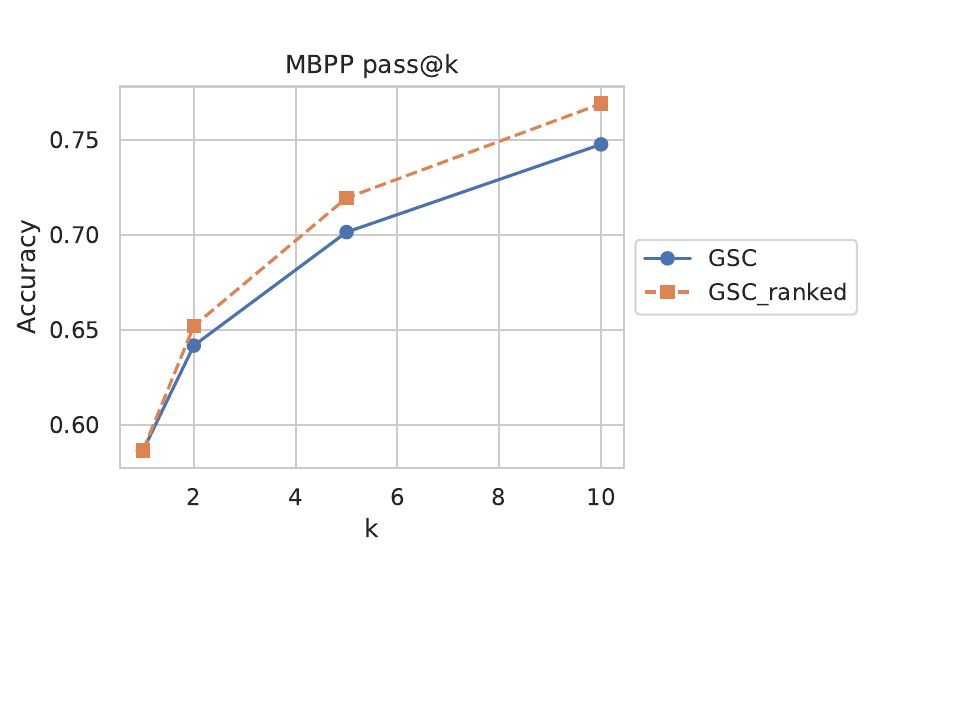}
  \end{minipage}
  \vspace{-0.4in}
  \begin{minipage}[b]{0.33\textwidth}
  \includegraphics[width=\textwidth]{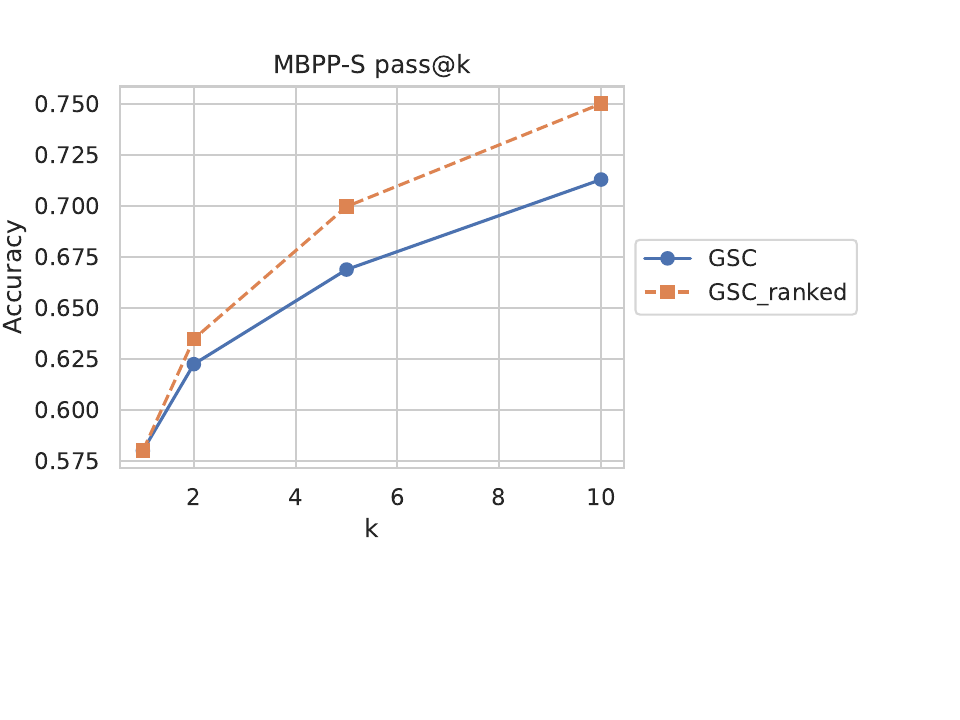}
  \end{minipage}
  \caption{\small{$pass@k$ for $k>1$ for HumanEval, MBPP, MBPP-S}}
  \label{fig:pass_at_k_eval}
\end{figure}

In Figure~\ref{fig:pass_at_k_eval} (Supplement), we show how the model performance changes as $k$ for $pass@k$ increases. We compare $GCS$ vs $GCS^{ranked}$. While the performance of $GCS$ declines quickly, $GCS^{ranked}$ maintains good performance even at larger values of $k$ for all code generation datasets.

\section{Related Work\label{background}}

\subsection{Auxiliary reranker}
In~\citet{mizumoto2016discriminative}, they use a perceptron based reranker to rerank model generated translations. SummaReranker~\citep{ravaut2022summareranker} use mixture of experts training to train their reranker to optimize for multiple automated evaluation metrics (like ROUGE or BLEU score) at once. PairReranker~\citep{jiang2022pairreranker} uses automated evaluation metrics to rank model generations and then select the top few best and worse and train a model to classify the better summary between a pair of summaries. All of the previous reranking methods however require training an auxiliary model.

\subsection{Code generation reranking}

There have also been multiple reranking proposals for code generation in particular. A unique characteristic of code (as oppposed to text) is that code can be executed. Thus several methods have tried to exploit that property for reranking. MBR-Exec~\citep{shi2022natural} and AlphaCode~\citep{li2022competition} both execute the generated codes on unit tests. They rank the different codes according to how many other codes are semantically equivalent to them (i.e. have the same results on the given unit tests). CodeT~\citep{chen2022codet} uses LLMs to generate both code and candidate unit tests. They then find sets of generated codes such that the product of the size of the set and the size of the \textit{unit test} set the codes agree on is maximized. More recently, Coder-Reviewer Ranker~\citep{zhang2022coder} applies the well known Maximum Mutual Information objective~\cite{li2015diversity} to code generating LLMs by using the strong few shot and zero prompting capabilities of LLMs to obtain the query likelihood.

\section{Conclusion\label{conclusion}}

We analyze the self-consistency method for problems that have fixed answers and develop a framework to extend it to open-ended generations. We establish connections between our framework and other code generation reranking functions and prove that if the optimal generation is present in our generation set, we can always recover it as well as prove bounds on how close we can get to the optimal generation under certain settings. 

Our simulated tests reveal our ability to consistently recover the best or close to best possible generation in the set. We introduce several lightweight similarity functions and show that they give strong and consistent improvements over state of the art baselines. Notably, our Unigram Consistency Score (UCS) function, the most minimal of our similarity functions, requires only access to raw generations to effectively rerank. We show that the UCS variants uniformly enhance the performance of code and text generation and are competitive with strong baselines like Coder Reviewer Reranker despite them needing a lot more compute resources as well as time. For code geneartion, we also leverage the fact that our reranking metric is based on pairwise similarity to improve performance for pass@$k$ for $k > 1$. Additionally, we conduct multiple variations on our primary experiments to ascertain the robustness and reliability of our performance.

\section{Broader Impact and Limitations}

As a paper that tries to improve the performance of Large Language Models (LLMs), it inherits the risk and rewards of LLMs in general. LLMs have shown themselves highly relevant and useful for a number of tasks but in particular code generation. Our method shows particularly strong improvements for that task and thus we hope will have a broad impact. Nevertheless, we did not evaluate our method on whether it increases its propensity to select biased or toxic generations which we leave to future work.

\bibliography{main}

\begin{thebibliography}{24}
\expandafter\ifx\csname natexlab\endcsname\relax\def\natexlab#1{#1}\fi

\bibitem[{Austin et~al.(2021)Austin, Odena, Nye, Bosma, Michalewski, Dohan,
  Jiang, Cai, Terry, Le et~al.}]{austin2021program}
Jacob Austin, Augustus Odena, Maxwell Nye, Maarten Bosma, Henryk Michalewski,
  David Dohan, Ellen Jiang, Carrie Cai, Michael Terry, Quoc Le, et~al. 2021.
\newblock Program synthesis with large language models.
\newblock \emph{arXiv preprint arXiv:2108.07732}.

\bibitem[{Bojar et~al.(2014)Bojar, Buck, Federmann, Haddow, Koehn, Leveling,
  Monz, Pecina, Post, Saint-Amand et~al.}]{bojar2014findings}
Ond{\v{r}}ej Bojar, Christian Buck, Christian Federmann, Barry Haddow, Philipp
  Koehn, Johannes Leveling, Christof Monz, Pavel Pecina, Matt Post, Herve
  Saint-Amand, et~al. 2014.
\newblock Findings of the 2014 workshop on statistical machine translation.
\newblock In \emph{Proceedings of the ninth workshop on statistical machine
  translation}, pages 12--58.

\bibitem[{Chen et~al.(2022)Chen, Zhang, Nguyen, Zan, Lin, Lou, and
  Chen}]{chen2022codet}
Bei Chen, Fengji Zhang, Anh Nguyen, Daoguang Zan, Zeqi Lin, Jian-Guang Lou, and
  Weizhu Chen. 2022.
\newblock Codet: Code generation with generated tests.
\newblock \emph{arXiv preprint arXiv:2207.10397}.

\bibitem[{Chen et~al.(2021)Chen, Tworek, Jun, Yuan, Pinto, Kaplan, Edwards,
  Burda, Joseph, Brockman et~al.}]{chen2021evaluating}
Mark Chen, Jerry Tworek, Heewoo Jun, Qiming Yuan, Henrique Ponde de~Oliveira
  Pinto, Jared Kaplan, Harri Edwards, Yuri Burda, Nicholas Joseph, Greg
  Brockman, et~al. 2021.
\newblock Evaluating large language models trained on code.
\newblock \emph{arXiv preprint arXiv:2107.03374}.

\bibitem[{Ho et~al.(2022)Ho, Schmid, and Yun}]{ho2022large}
Namgyu Ho, Laura Schmid, and Se-Young Yun. 2022.
\newblock Large language models are reasoning teachers.
\newblock \emph{arXiv preprint arXiv:2212.10071}.

\bibitem[{Holtzman et~al.(2019)Holtzman, Buys, Du, Forbes, and
  Choi}]{holtzman2019curious}
Ari Holtzman, Jan Buys, Li~Du, Maxwell Forbes, and Yejin Choi. 2019.
\newblock The curious case of neural text degeneration.
\newblock \emph{arXiv preprint arXiv:1904.09751}.

\bibitem[{Hsieh et~al.(2023)Hsieh, Li, Yeh, Nakhost, Fujii, Ratner, Krishna,
  Lee, and Pfister}]{hsieh2023distilling}
Cheng-Yu Hsieh, Chun-Liang Li, Chih-Kuan Yeh, Hootan Nakhost, Yasuhisa Fujii,
  Alexander Ratner, Ranjay Krishna, Chen-Yu Lee, and Tomas Pfister. 2023.
\newblock Distilling step-by-step! outperforming larger language models with
  less training data and smaller model sizes.
\newblock \emph{arXiv preprint arXiv:2305.02301}.

\bibitem[{Jiang et~al.(2022{\natexlab{a}})Jiang, Welleck, Zhou, Li, Liu,
  Jamnik, Lacroix, Wu, and Lample}]{jiang2022draft}
Albert~Q Jiang, Sean Welleck, Jin~Peng Zhou, Wenda Li, Jiacheng Liu, Mateja
  Jamnik, Timoth{\'e}e Lacroix, Yuhuai Wu, and Guillaume Lample.
  2022{\natexlab{a}}.
\newblock Draft, sketch, and prove: Guiding formal theorem provers with
  informal proofs.
\newblock \emph{arXiv preprint arXiv:2210.12283}.

\bibitem[{Jiang et~al.(2022{\natexlab{b}})Jiang, Lin, and
  Ren}]{jiang2022pairreranker}
Dongfu Jiang, Bill~Yuchen Lin, and Xiang Ren. 2022{\natexlab{b}}.
\newblock Pairreranker: Pairwise reranking for natural language generation.
\newblock \emph{arXiv preprint arXiv:2212.10555}.

\bibitem[{Li et~al.(2015)Li, Galley, Brockett, Gao, and
  Dolan}]{li2015diversity}
Jiwei Li, Michel Galley, Chris Brockett, Jianfeng Gao, and Bill Dolan. 2015.
\newblock A diversity-promoting objective function for neural conversation
  models.
\newblock \emph{arXiv preprint arXiv:1510.03055}.

\bibitem[{Li et~al.(2022)Li, Choi, Chung, Kushman, Schrittwieser, Leblond,
  Eccles, Keeling, Gimeno, Dal~Lago et~al.}]{li2022competition}
Yujia Li, David Choi, Junyoung Chung, Nate Kushman, Julian Schrittwieser,
  R{\'e}mi Leblond, Tom Eccles, James Keeling, Felix Gimeno, Agustin Dal~Lago,
  et~al. 2022.
\newblock Competition-level code generation with alphacode.
\newblock \emph{Science}, 378(6624):1092--1097.

\bibitem[{Liu and Liu(2021)}]{liu2021simcls}
Yixin Liu and Pengfei Liu. 2021.
\newblock Simcls: A simple framework for contrastive learning of abstractive
  summarization.
\newblock \emph{arXiv preprint arXiv:2106.01890}.

\bibitem[{Mizumoto and Matsumoto(2016)}]{mizumoto2016discriminative}
Tomoya Mizumoto and Yuji Matsumoto. 2016.
\newblock Discriminative reranking for grammatical error correction with
  statistical machine translation.
\newblock In \emph{Proceedings of the 2016 Conference of the North American
  Chapter of the Association for Computational Linguistics: Human Language
  Technologies}, pages 1133--1138.

\bibitem[{Narayan et~al.(2018)Narayan, Cohen, and Lapata}]{narayan2018don}
Shashi Narayan, Shay~B Cohen, and Mirella Lapata. 2018.
\newblock Don't give me the details, just the summary! topic-aware
  convolutional neural networks for extreme summarization.
\newblock \emph{arXiv preprint arXiv:1808.08745}.

\bibitem[{Ouyang et~al.(2022)Ouyang, Wu, Jiang, Almeida, Wainwright, Mishkin,
  Zhang, Agarwal, Slama, Ray et~al.}]{ouyang2022training}
Long Ouyang, Jeffrey Wu, Xu~Jiang, Diogo Almeida, Carroll Wainwright, Pamela
  Mishkin, Chong Zhang, Sandhini Agarwal, Katarina Slama, Alex Ray, et~al.
  2022.
\newblock Training language models to follow instructions with human feedback.
\newblock \emph{Advances in Neural Information Processing Systems},
  35:27730--27744.

\bibitem[{Polu et~al.(2022)Polu, Han, Zheng, Baksys, Babuschkin, and
  Sutskever}]{polu2022formal}
Stanislas Polu, Jesse~Michael Han, Kunhao Zheng, Mantas Baksys, Igor
  Babuschkin, and Ilya Sutskever. 2022.
\newblock Formal mathematics statement curriculum learning.
\newblock \emph{arXiv preprint arXiv:2202.01344}.

\bibitem[{Ravaut et~al.(2022)Ravaut, Joty, and Chen}]{ravaut2022summareranker}
Mathieu Ravaut, Shafiq Joty, and Nancy~F Chen. 2022.
\newblock Summareranker: A multi-task mixture-of-experts re-ranking framework
  for abstractive summarization.
\newblock \emph{arXiv preprint arXiv:2203.06569}.

\bibitem[{Shi et~al.(2022)Shi, Fried, Ghazvininejad, Zettlemoyer, and
  Wang}]{shi2022natural}
Freda Shi, Daniel Fried, Marjan Ghazvininejad, Luke Zettlemoyer, and Sida~I
  Wang. 2022.
\newblock Natural language to code translation with execution.
\newblock \emph{arXiv preprint arXiv:2204.11454}.

\bibitem[{Uesato et~al.(2022)Uesato, Kushman, Kumar, Song, Siegel, Wang,
  Creswell, Irving, and Higgins}]{uesato2022solving}
Jonathan Uesato, Nate Kushman, Ramana Kumar, Francis Song, Noah Siegel, Lisa
  Wang, Antonia Creswell, Geoffrey Irving, and Irina Higgins. 2022.
\newblock Solving math word problems with process-and outcome-based feedback.
\newblock \emph{arXiv preprint arXiv:2211.14275}.

\bibitem[{Wainwright(2019)}]{wainwright2019high}
Martin~J Wainwright. 2019.
\newblock \emph{High-dimensional statistics: A non-asymptotic viewpoint},
  volume~48.
\newblock Cambridge university press.

\bibitem[{Wang et~al.(2022)Wang, Wei, Schuurmans, Le, Chi, and
  Zhou}]{wang2022self}
Xuezhi Wang, Jason Wei, Dale Schuurmans, Quoc Le, Ed~Chi, and Denny Zhou. 2022.
\newblock Self-consistency improves chain of thought reasoning in language
  models.
\newblock \emph{arXiv preprint arXiv:2203.11171}.

\bibitem[{Welleck et~al.(2019)Welleck, Kulikov, Roller, Dinan, Cho, and
  Weston}]{welleck2019neural}
Sean Welleck, Ilia Kulikov, Stephen Roller, Emily Dinan, Kyunghyun Cho, and
  Jason Weston. 2019.
\newblock Neural text generation with unlikelihood training.
\newblock \emph{arXiv preprint arXiv:1908.04319}.

\bibitem[{Wu et~al.(2022)Wu, Jiang, Li, Rabe, Staats, Jamnik, and
  Szegedy}]{wu2022autoformalization}
Yuhuai Wu, Albert~Qiaochu Jiang, Wenda Li, Markus Rabe, Charles Staats, Mateja
  Jamnik, and Christian Szegedy. 2022.
\newblock Autoformalization with large language models.
\newblock \emph{Advances in Neural Information Processing Systems},
  35:32353--32368.

\bibitem[{Zhang et~al.(2022)Zhang, Yu, Hashimoto, Lewis, Yih, Fried, and
  Wang}]{zhang2022coder}
Tianyi Zhang, Tao Yu, Tatsunori~B Hashimoto, Mike Lewis, Wen-tau Yih, Daniel
  Fried, and Sida~I Wang. 2022.
\newblock Coder reviewer reranking for code generation.
\newblock \emph{arXiv preprint arXiv:2211.16490}.

\end{thebibliography}

\clearpage

\section*{Supplementary Material}

\renewcommand\thesubsection{\Alph{subsection}}

\subsection{Proofs}

\subsubsection{Proof of Theorem 2.1}

\begin{proof}
This is true by definition for $k=1$. For $k>1$, let us assume that the number of categories $L = 3$. If the best generation $g$ agrees with $\mathbf{v}$ on only one of the elements, then wlog, let that be the 1st one. Then the agreement score is $(p_1 + p'_2)/2$ where $p'_2 < p_2$. Let the agreement score for a generation $g'$ that does not agree at all with $\mathbf{v}$ be $(p'_1 + p''_2)/2$. However if for example $p_1 = 0.34, p'_1 = 0.32, p'_2 = 0.01, p''_2 = 0.32$, then $g'$ will be selected over $g$.
\end{proof}

\subsubsection{Proof of Theorem 2.2}

\begin{proof}
It is true by assumption for $k=1$. Assume it is true for $k=t$. Then that means that given the self consistency assumption that $a_t(\mathbf{u}_b, \mathbf{v})$ is the highest possible where $a_t$ is the agreement until $k=t$. Then for $t+1$, we know that $\sum_{i \neq b} \mathbb{I}(\mathbf{u}_b^{t+1} = \mathbf{u}_i^{t+1}$ is the highest (again by self-consistency assumption). Thus $a_{t+1}$ is also the highest proving the theorem.
\end{proof}

\subsubsection{Proof of Theorem 2.3}

Formally, let $\mathbf{u}^j_i \sim Bernoulli(p_j)$. Let $b = \arg \max_i \sum^j p_j\cdot \mathbf{u}^j_i + (1-p_j)\cdot (1-\mathbf{u}^j_i) = \arg \max_i \sum^j \mathbf{u}^j_i \cdot (2p_j-1)$ (i.e. the sequence selected by our method). Then we want a bound on $\mathbb{E}[\sum_j^k \mathbf{u}_b^j]$.

\begin{proof}
Let $q_i = \sum_j \mathbf{u}^j_i$. As all are iid, $\mathbb{E}[q_i] = \sum_j p_j$. We can upper bound this by upper bounding $\mathbb{E}[\max_i q_i]$. Note that $\mathbf{u}^j_i$ is subgaussian with parameter 1/2 as it's bounded in $[0, 1]$. Thus $q_i$ is subgaussian with parameter $\sqrt{k}/2$. Thus $\mathbb{E}[\max{q_i - \mathbb{E}[q_j]}] \leq \sqrt{\frac{k \log{n}}{2}} \implies \mathbb{E}[\max{q_i}] \leq \sum_i p_i + \sqrt{\frac{k \log{n}}{2}}$ where $n$ is the number of generations~\cite{wainwright2019high}

\end{proof}

\subsection{Simulation results}

\begin{figure}[ht]
\begin{adjustwidth}{-2cm}{-2cm}
\captionsetup{justification=centering}
  \centering
  \begin{minipage}[b]{0.45\textwidth}
  \includegraphics[width=\textwidth]{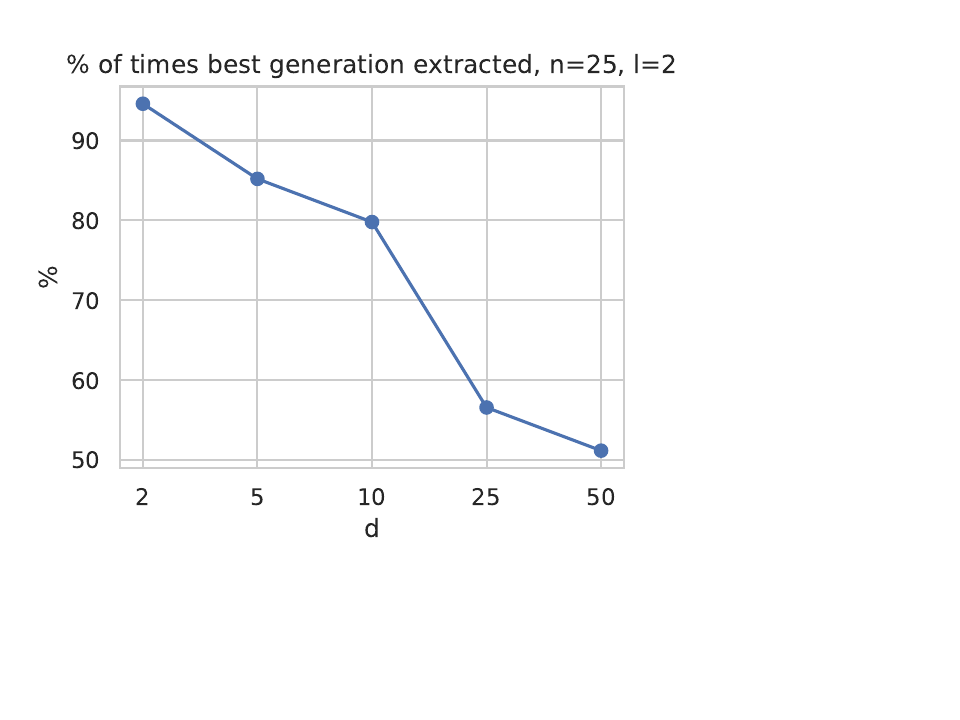}
  \end{minipage}
  \hspace{-0.6in}
  \begin{minipage}[b]{0.45\textwidth}
  \includegraphics[width=\textwidth]{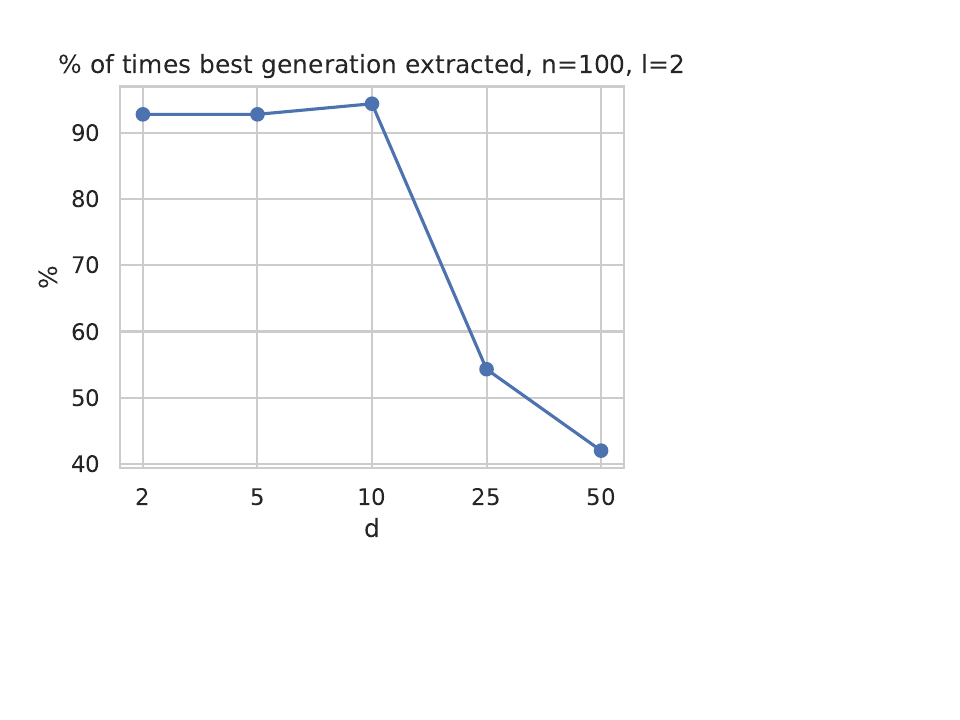}
  \end{minipage}
  \hspace{-0.6in}
  \begin{minipage}[b]{0.45\textwidth}
  \includegraphics[width=\textwidth]{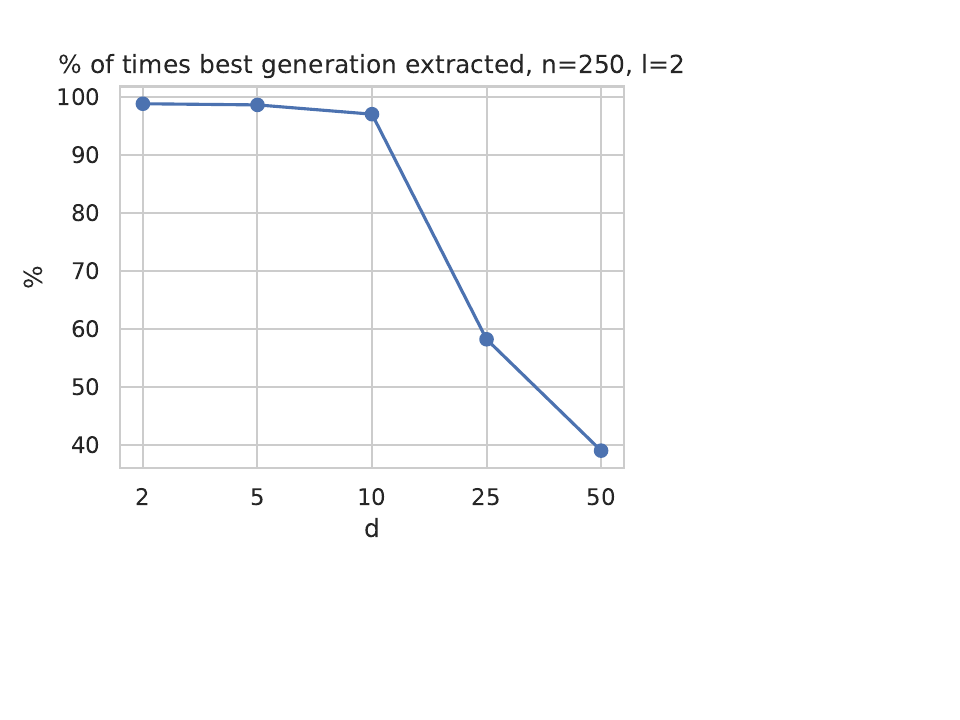}
  \end{minipage}
  \begin{minipage}[b]{0.45\textwidth}
  \includegraphics[width=\textwidth]{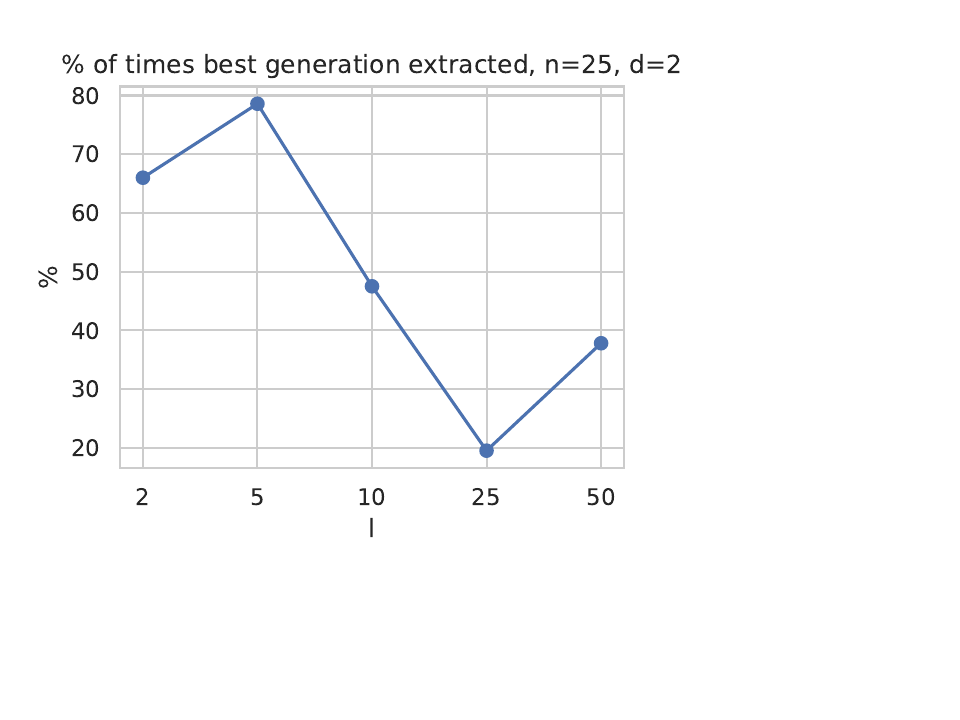}
  \end{minipage}
  \hspace{-0.6in}
  \begin{minipage}[b]{0.45\textwidth}
  \includegraphics[width=\textwidth]{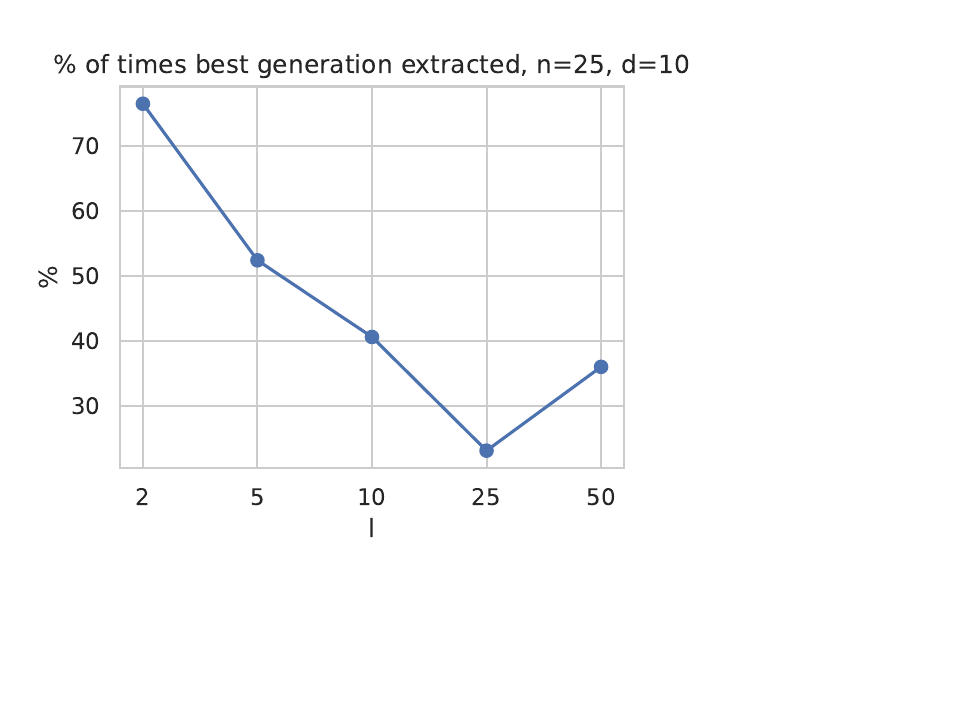}
  \end{minipage}
  \hspace{-0.6in}
  \begin{minipage}[b]{0.45\textwidth}
  \includegraphics[width=\textwidth]{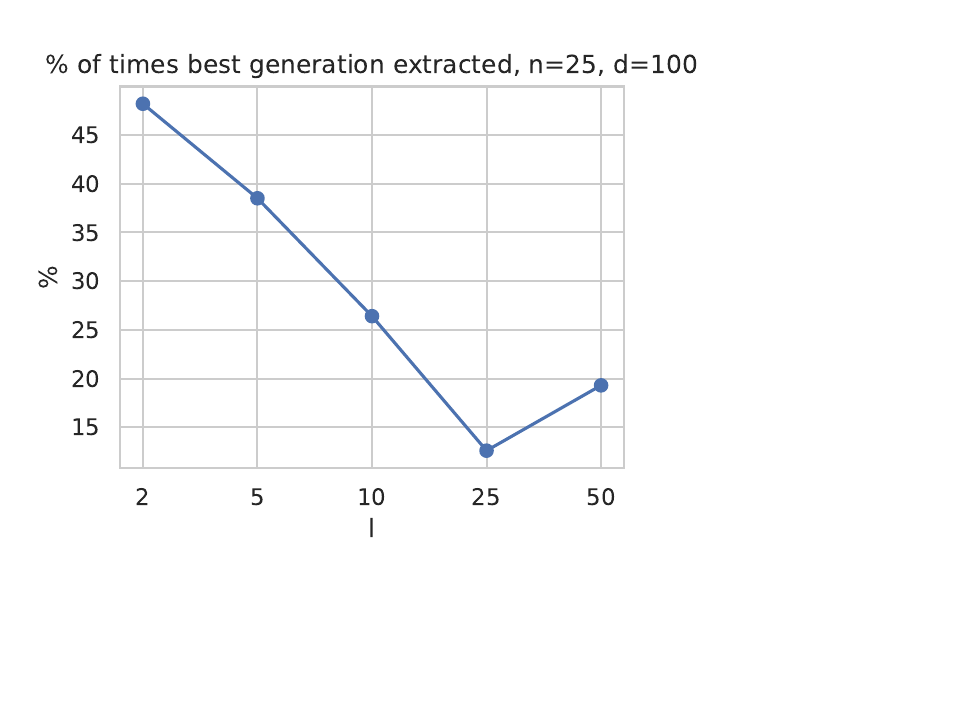}
  \end{minipage}
  \begin{minipage}[b]{0.45\textwidth}
  \includegraphics[width=\textwidth]{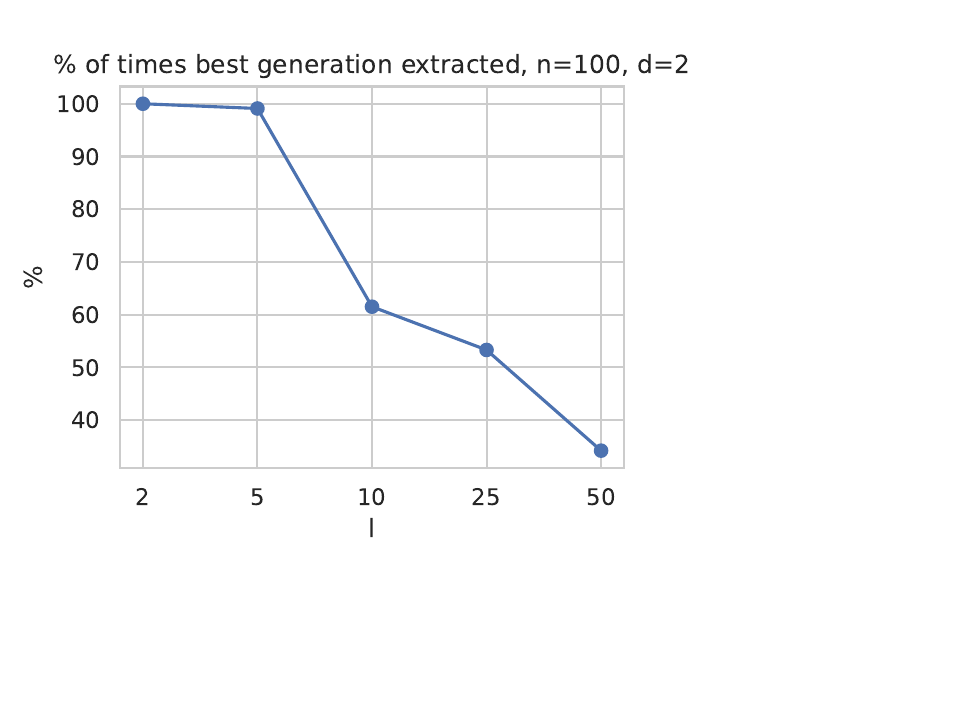}
  \end{minipage}
  \hspace{-0.6in}
  \begin{minipage}[b]{0.45\textwidth}
  \includegraphics[width=\textwidth]{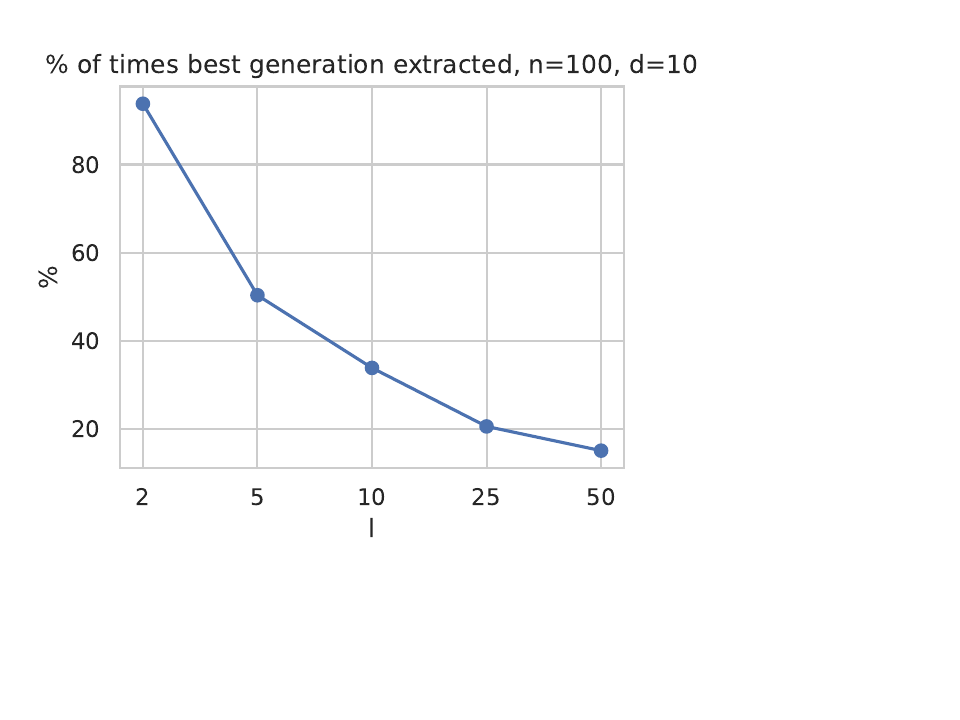}
  \end{minipage}
  \hspace{-0.6in}
  \begin{minipage}[b]{0.45\textwidth}
  \includegraphics[width=\textwidth]{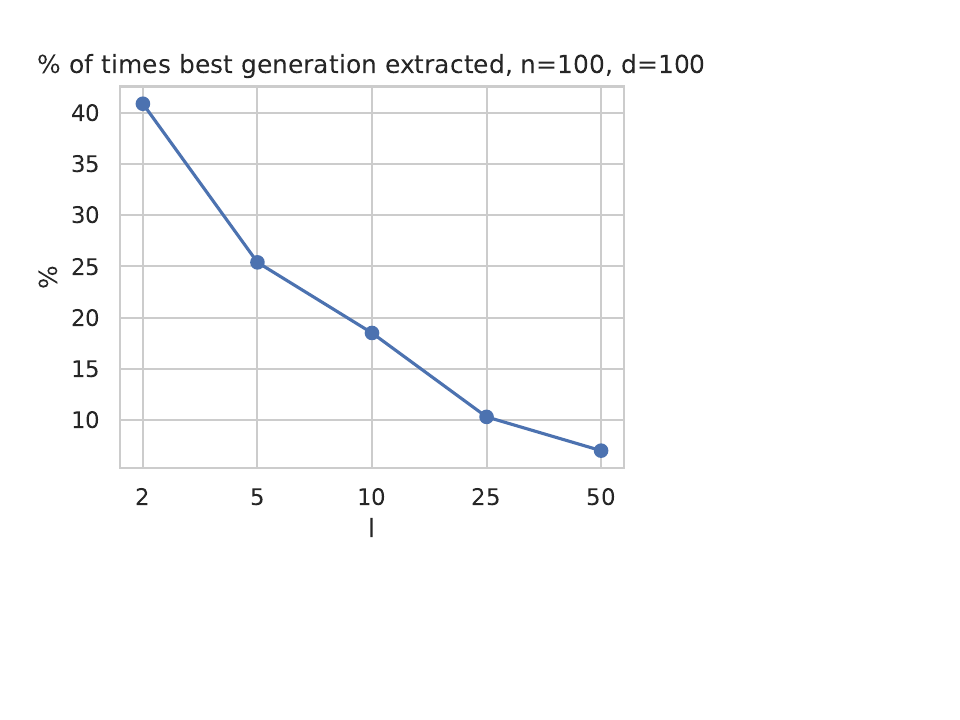}
  \end{minipage}
  \begin{minipage}[b]{0.45\textwidth}
  \includegraphics[width=\textwidth]{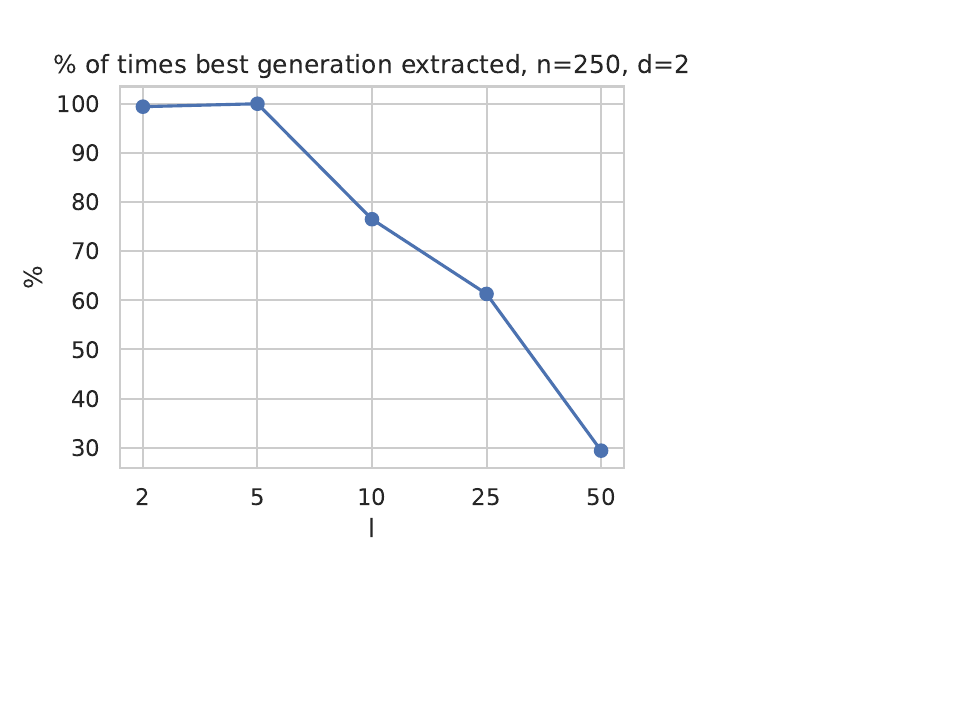}
  \end{minipage}
  \hspace{-0.6in}
  \begin{minipage}[b]{0.45\textwidth}
  \includegraphics[width=\textwidth]{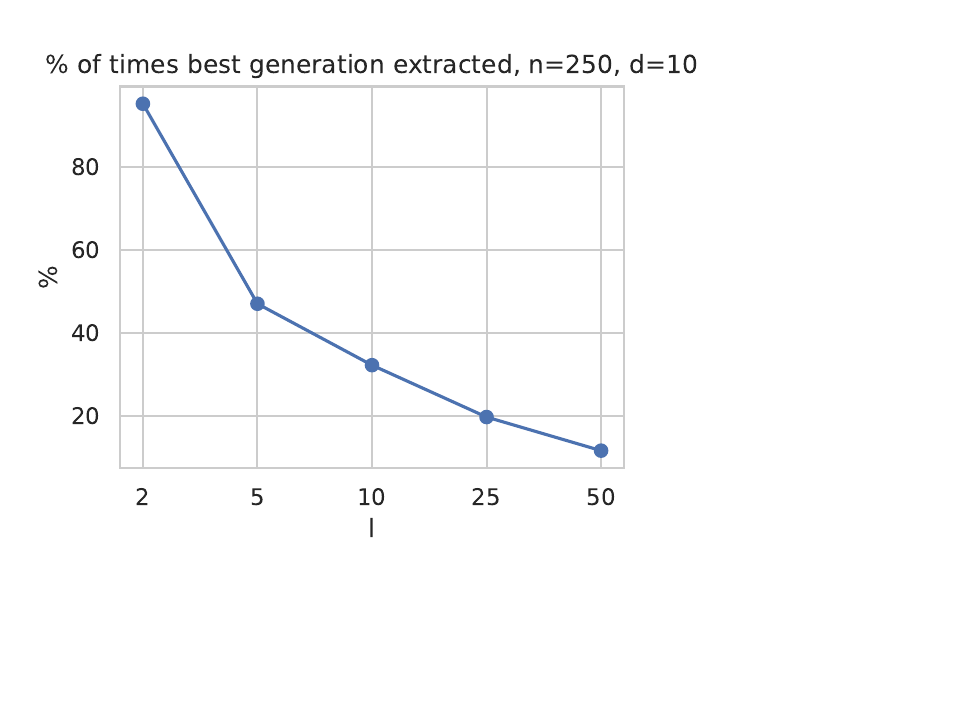}
  \end{minipage}
  \hspace{-0.6in}
  \begin{minipage}[b]{0.45\textwidth}
  \includegraphics[width=\textwidth]{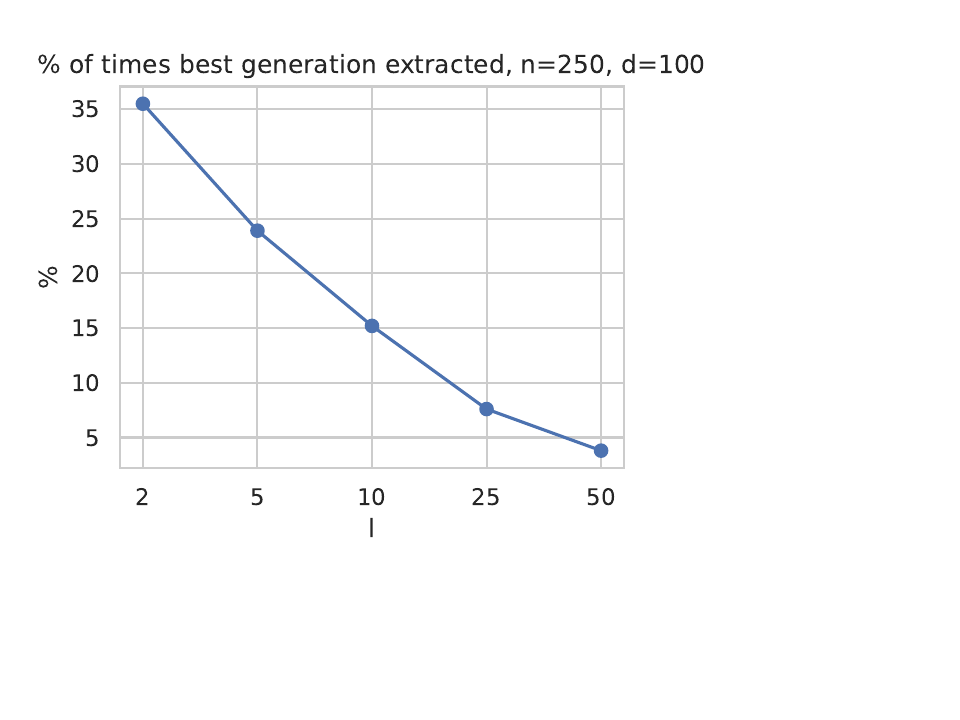}
  \end{minipage}
  \end{adjustwidth}
  \caption{The above figures show what percentage of the time we are able to retrieve the best generation out of the set of generations that we have}
  \label{fig:top1}
\end{figure}

\begin{figure}[ht]
\begin{adjustwidth}{-2cm}{-2cm}
\captionsetup{justification=centering}
  \centering
  \begin{minipage}[b]{0.45\textwidth}
  \includegraphics[width=\textwidth]{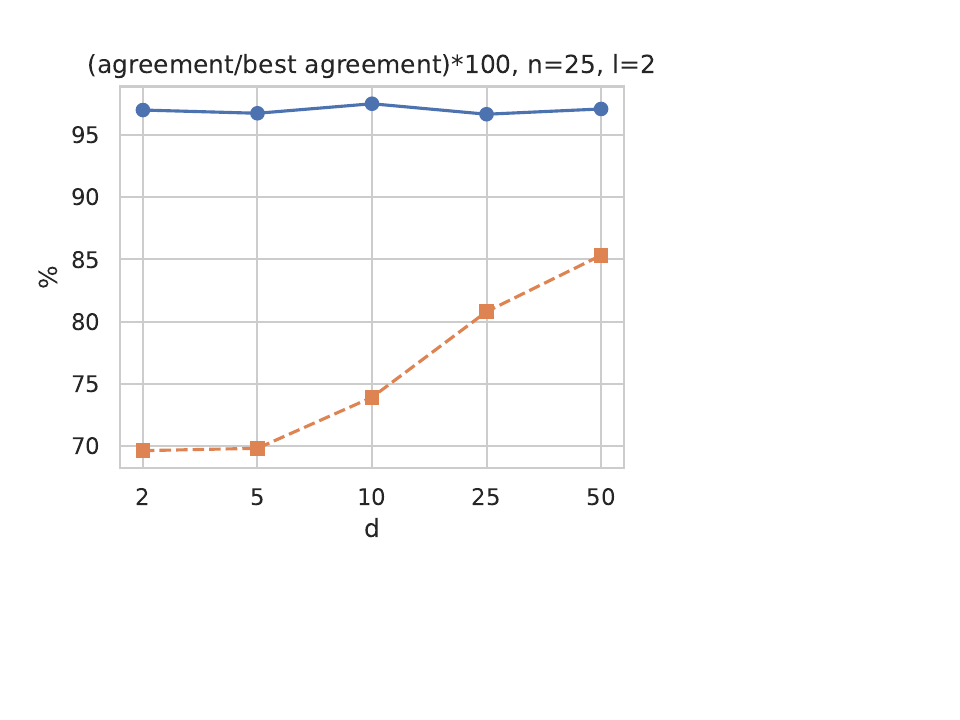}
  \end{minipage}
  \hspace{-0.6in}
  \begin{minipage}[b]{0.45\textwidth}
  \includegraphics[width=\textwidth]{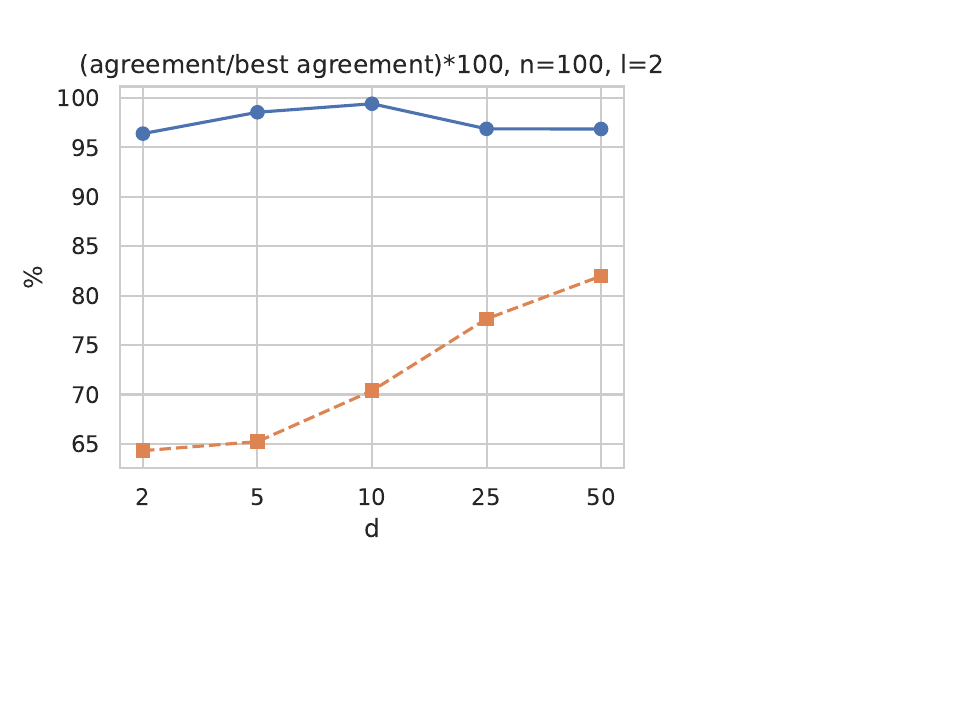}
  \end{minipage}
  \hspace{-0.6in}
  \begin{minipage}[b]{0.45\textwidth}
  \includegraphics[width=\textwidth]{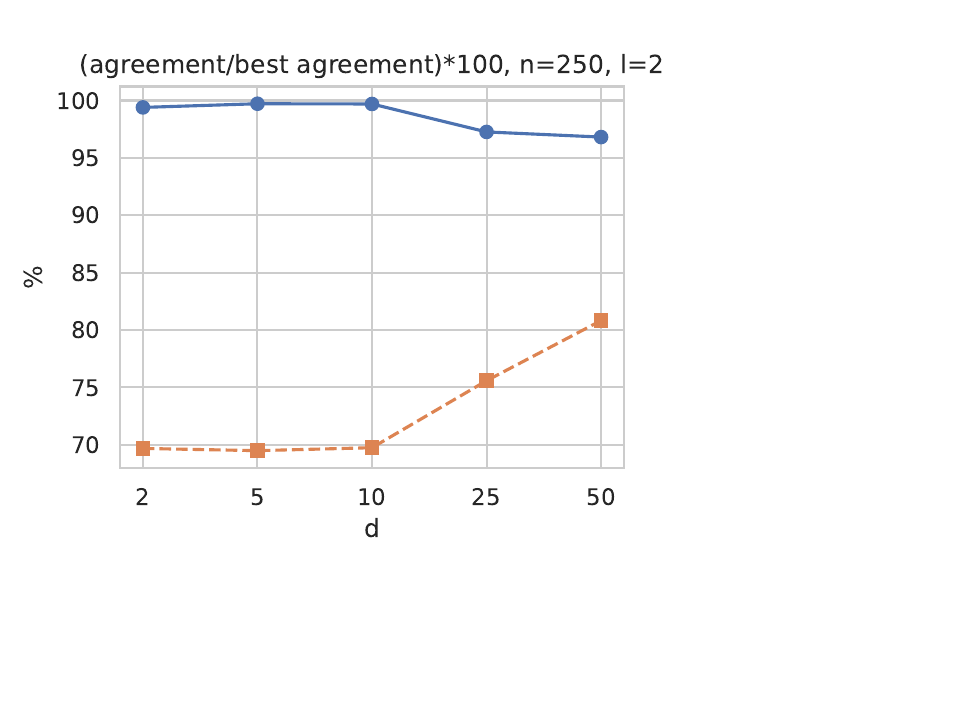}
  \end{minipage}
  \begin{minipage}[b]{0.45\textwidth}
  \includegraphics[width=\textwidth]{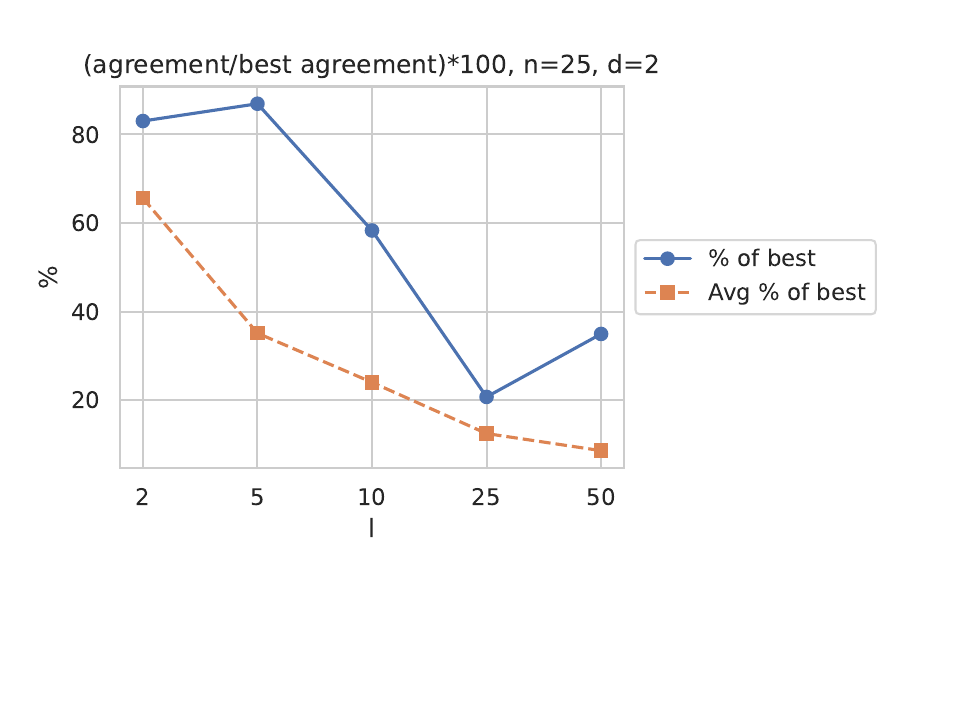}
  \end{minipage}
  \hspace{-0.6in}
  \begin{minipage}[b]{0.45\textwidth}
  \includegraphics[width=\textwidth]{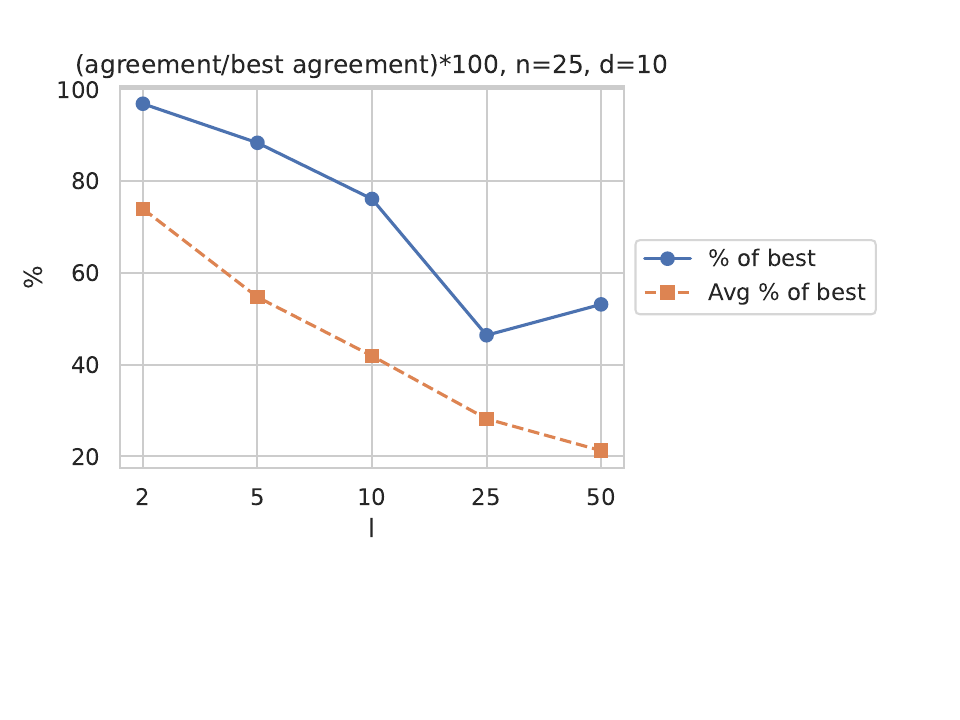}
  \end{minipage}
  \hspace{-0.6in}
  \begin{minipage}[b]{0.45\textwidth}
  \includegraphics[width=\textwidth]{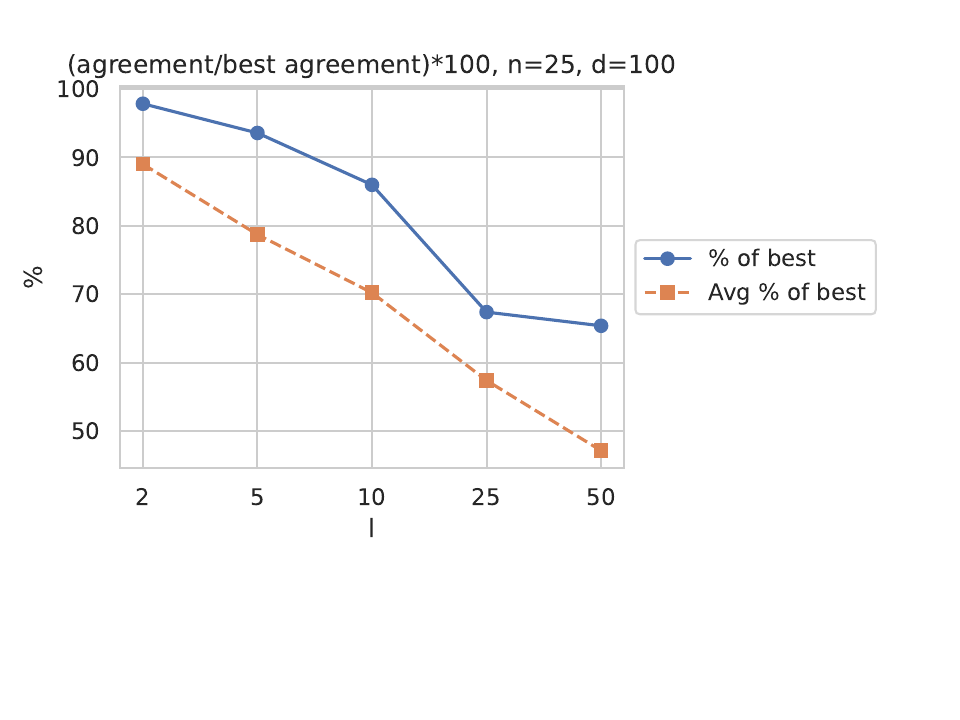}
  \end{minipage}
  \begin{minipage}[b]{0.45\textwidth}
  \includegraphics[width=\textwidth]{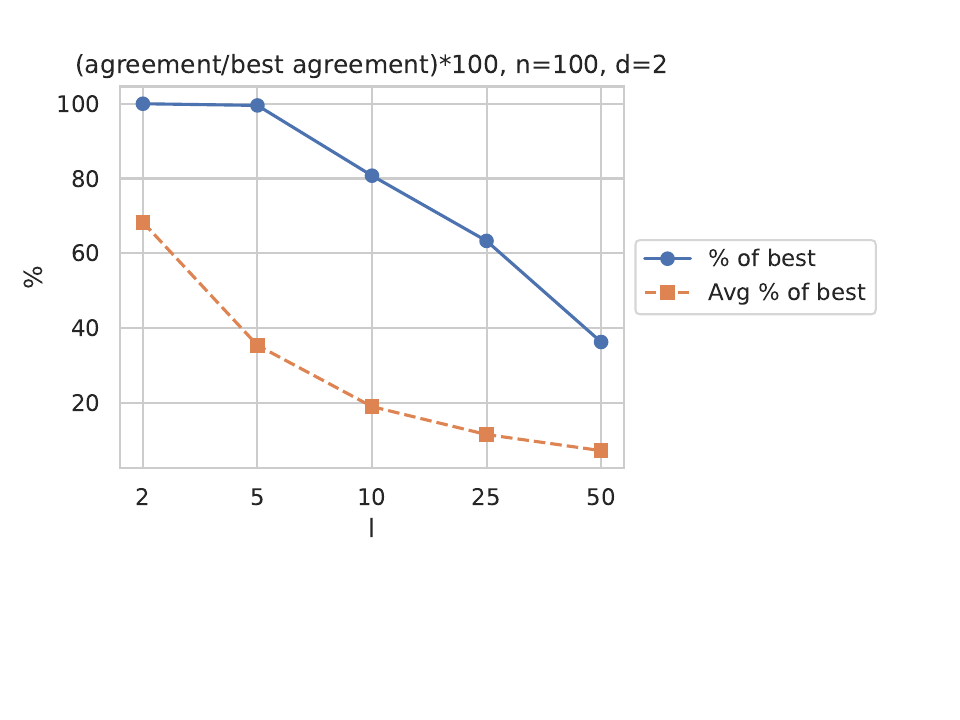}
  \end{minipage}
  \hspace{-0.6in}
  \begin{minipage}[b]{0.45\textwidth}
  \includegraphics[width=\textwidth]{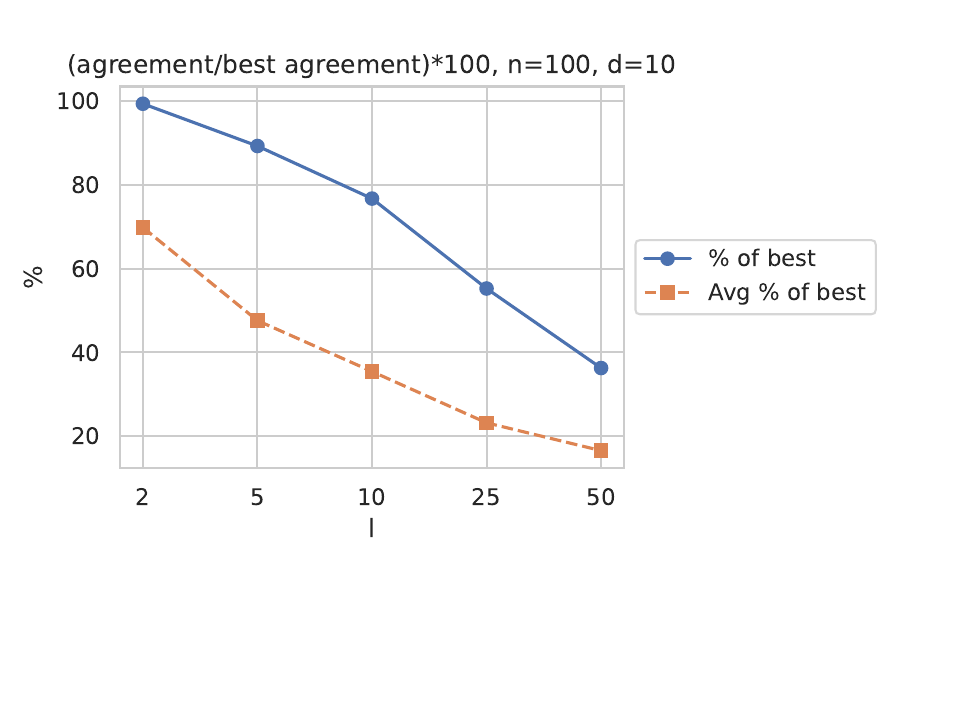}
  \end{minipage}
  \hspace{-0.6in}
  \begin{minipage}[b]{0.45\textwidth}
  \includegraphics[width=\textwidth]{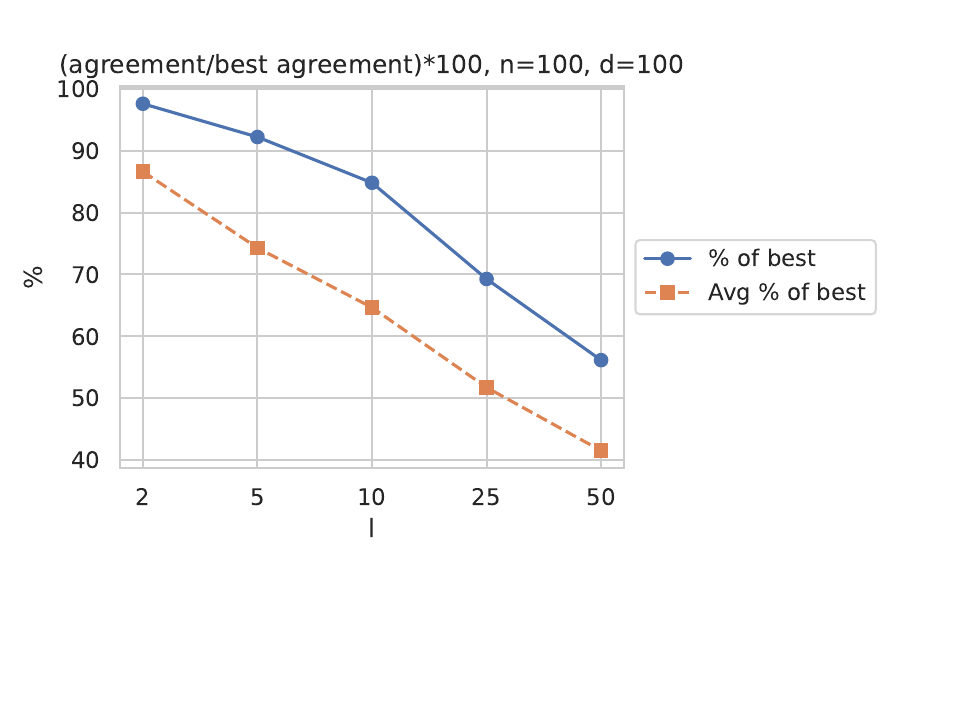}
  \end{minipage}
  \begin{minipage}[b]{0.45\textwidth}
  \includegraphics[width=\textwidth]{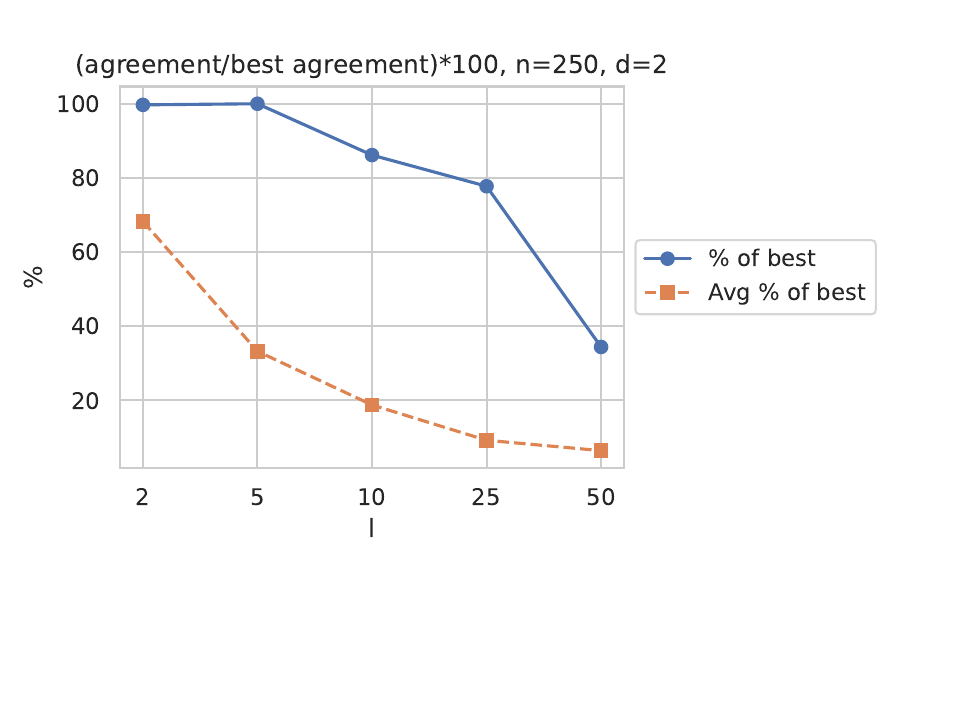}
  \end{minipage}
  \hspace{-0.6in}
  \begin{minipage}[b]{0.45\textwidth}
  \includegraphics[width=\textwidth]{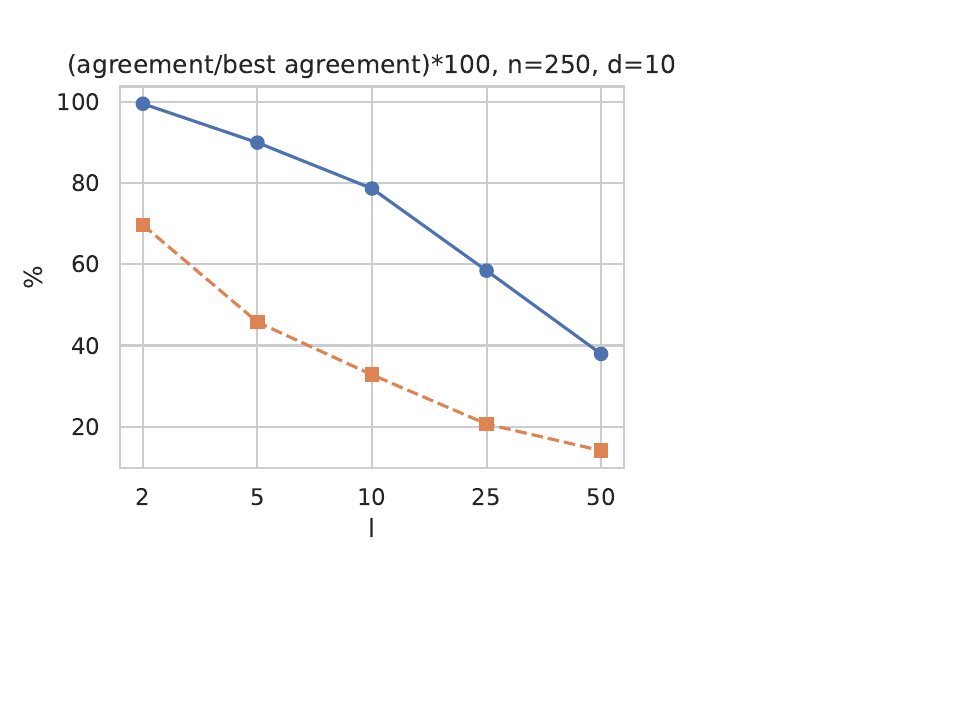}
  \end{minipage}
  \hspace{-0.6in}
  \begin{minipage}[b]{0.45\textwidth}
  \includegraphics[width=\textwidth]{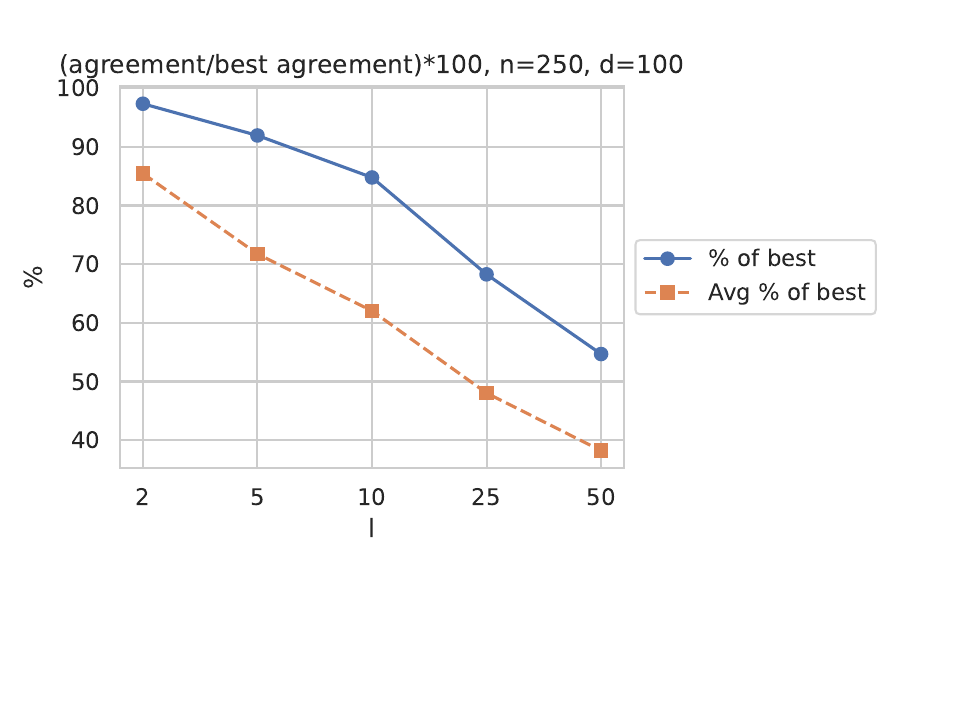}
  \end{minipage}
  \end{adjustwidth}
  \caption{The above figures show what \% the best generation as per the highest fractional agreement heuristic and a randomly selected generation agree with the best generation of the set}
  \label{fig:perbest}
\end{figure}

We setup our simulation as follows. Let $d$ be the number of predicates, $n$ the number of generations, and $l$ the number of categories. Then for each predicate, we uniformly at random sample a categorical distribution and then generate $\mathbf{u}_i$ from that distribution. We then apply our criterion of picking the $\mathbf{u}_b$ that has the highest average fractional agreement with all other $\mathbf{u}_i$ and measure (1) the \% of times we are able to retrieve the generation that has the best agreement with $\mathbf{v}$ (2) the \% agreement $\mathbf{u}_b$ has with the best possible generation out of the set. We vary $d, l$ between $2$ and $50$, and $n$ between $25$ and $250$. All our results are based on 1000 samples. The results are in Figures~\ref{fig:top1} and~\ref{fig:perbest}. 

For the first metric, we are able to retrieve the best generation a very high fraction of the time when $l$ is $<5$ even when $d$ goes to higher values. Even when $l$ is larger, we are still able to retrieve the best generation a non-trivial fraction of times -- and notably our performance does not degrade much as $n$ goes from $25$ to $250$.

Turning our attention to the second metric, we are able to consistently get a generation close to the best generation. This is especially true for small $l$ where even when $d$ increases to large values, we are able to get close to 100\% agreement with the best generation. Even at high values of $l$ however, we get relatively good agreement with the best generation -- especially compared to picking a random generation -- a heuristic we consistently beat.

\begin{table*}
\centering
\fontsize{8.25pt}{8.25pt}\selectfont
\setlength\tabcolsep{6.25pt} 
\scalebox{0.9}{
\begin{tabular}{p{2cm}cccccc}
\toprule
 & \textbf{Medoid} & \textbf{Mean-logp} & \textbf{UCS} & \textbf{WUCS} & \textbf{Consensus-WUCS}\\
\midrule
 & \textbf{HumanEval}\\
\textbf{Codex002} & 0.515 & 0.604 & 0.615 & \second{0.630} & \first{0.633}\\
\textbf{Codex001} & 0.432 & 0.484 & 0.488 & \second{0.507} & \first{0.517}\\
\textbf{Code-Cushman} & 0.4 & 0.428 & 0.434 & \second{0.451} & \first{0.454}\\
\textbf{Llama-13B} & 0.231 & 0.221 & 0.242 & \second{0.248} & \first{0.25}\\
\textbf{Llama-30B} & 0.29 & 0.286 & 0.324 & \first{0.327} & \first{0.327}\\
\midrule
 & \textbf{MBPP-S}\\
\textbf{Codex002} & 0.64 & 0.626 & \first{0.67} & 0.643 & \second{0.647}\\
\textbf{Codex001} & 0.594 & 0.575 & 0.594 & \second{0.599} & \first{0.605}\\
\textbf{Code-Cushman} & 0.527 & 0.521 & 0.531 & \second{0.541} & \first{0.549}\\
\textbf{Llama-13B} & \first{0.355} & 0.331 & 0.340  & 0.344 & \second{0.347} \\
\textbf{Llama-30B} & 0.425 & 0.408 & 0.337 & \second{0.436} & \first{0.438}\\
\midrule
 & \textbf{MBPP}\\
\textbf{Codex002} & 0.631 & 0.549 & 0.651 & \second{0.655} & \first{0.659}\\
\textbf{Codex001} & 0.574 & 0.58 & 0.587 & \second{0.596} & \first{0.598}\\
\textbf{Code-Cushman} & 0.435 & 0.29 & 0.479 & \second{0.494} & \first{0.503}\\
\textbf{Llama-13B} & 0.269 & 0.3 & 0.261 & \first{0.305} & \second{0.304}\\
\textbf{Llama-30B} & 0.346 & 0.332 & 0.351 & \second{0.358} & \first{0.359}\\
\\ 
\bottomrule
\end{tabular}
}
\caption{\small{Mean reciprocal rank of generations for HumanEval, MBPP, MBBP-S}. Best results are colored in \first{first}, \second{second}.}\label{table:code_gen_mrr}
\end{table*}

\subsection{Experimental baselines}

As mentioned earlier, we could not obtain Codex-001 and Codex-Cushman results on Xsum and MiniF2F due to the unexpected API shutdown. For the BLEU and Rouge-2 metrics, we report the values divided by 100. In terms of our baselines, we have

\begin{enumerate}
    \item \textbf{Random selection} - we randomly select a generation from the set of generations
    \item \textbf{Ranking by mean log probability} - we take the average log probability across the tokens in the generation and select the generation with the highest mean log probability
    \item \textbf{Ranking using Medoid} - we take the generation with the lowest mean distance to all other generations in our confidence weighted unigram space as used in WUCS.
    \item \textbf{Coder Reviewer Ranker} - This method has two variants -- Normalized Reviewer (NR), and Normalized Coder Reviewer (NCR). NR computes the mean per token $\log{p(x|y)}$, where $y$ is the generation and $x$ is the prompt, and then ranks based on this metric. On the other hand, NCR merges the mean log probability ranking with NR, ranking according to $\log{p(x|y)} + \log{p(y|x)}$. As the state of the art in code reranking, these methods represent a strong baseline.
\end{enumerate}

\subsection{Comparison with Coder-Reviewer Ranker}

The comparison with the Code Reviewer Ranker baseline, specifically with the Normalized Reviewer (NR) and Normalized Coder-Reviewer (NCR) variants, is in Table~\ref{table:crr}. As the state of the art in code reranking, these methods represent a strong baseline.
Our results demonstrate that the WUCS and Consensus-WUCS methods are highly competitive. Consensus-WUCS consistently outperforms NR and often surpasses NCR as well, despite the fact that NR and NCR require a second forward pass, which doubles the inference cost and adds latency overhead. 

In the HumanEval dataset, Consensus-WUCS yields the highest accuracy for the Llama-13B and Llama-30B models. Similarly, in the MBPP-S dataset, Consensus-WUCS delivers superior performance for the Llama-13B and Llama-30B models, and closely matches the NCR for Codex models. In the MBPP dataset, the Consensus-WUCS method ranks as the best for Code-Cushman, Llama-13B, and Llama-30B models.

Notably in 40\% of the experiments (6 out of 15), Consensus-WUCS outperforms all other methods, including the highly competitive NCR. Furthermore, Consensus-WUCS ranks second in 8 out of the 15 experiments, reinforcing its strong performance across diverse models and datasets. 

Our results present evidence of the effectiveness of WUCS and Consensus-WUCS, which hold their own against much more heavyweight state-of-the-art methods and frequently deliver superior performance.

\begin{table*}
\centering 
\fontsize{8.25pt}{8.25pt}\selectfont
\setlength\tabcolsep{6.25pt} 
\scalebox{0.9}{
\begin{tabular}{p{4cm}ccccc}
\toprule
 & \textbf{WUCS} & \textbf{Consensus-WUCS} & \textbf{N. Reviewer} & \textbf{N. Coder-Reviewer}\\
\midrule
 & \textbf{HumanEval}\\
\textbf{Codex002} & 0.558 & \second{0.568} & 0.524 & \first{0.576}\\
\textbf{Codex001} & 0.426 & \second{0.445} & 0.42 & \first{0.482}\\
\textbf{Code-Cushman} & 0.373 & \second{0.381} & 0.358 & \first{0.385}\\
\textbf{Llama-13B} & \second{0.187} & \first{0.192} & 0.164 & 0.181\\
\textbf{Llama-30B} & \second{0.263} & \first{0.267} & 0.219 & 0.241\\
\midrule
 & \textbf{MBPP-S}\\
\textbf{Codex002} & 0.58 & \second{0.589} & 0.559 & \first{0.595}\\
\textbf{Codex001} & 0.535 & \second{0.546} & 0.509 & \first{0.55}\\
\textbf{Code-Cushman} & 0.472 & \second{0.488} & 0.455 & \first{0.512}\\
\textbf{Llama-13B} & \second{0.266} & \first{0.277} & 0.228 & \second{0.266}\\
\textbf{Llama-30B} & \second{0.363} & \first{0.373} & 0.302 & 0.325\\
\midrule
 & \textbf{MBPP}\\
\textbf{Codex002} & 0.587 & \second{0.594} & \first{0.631} & 0.592\\
\textbf{Codex001} & 0.52 & 0.525 & \second{0.532} & \first{0.545}\\
\textbf{Code-Cushman} & \second{0.405} & \first{0.42} & 0.398 & 0.339\\
\textbf{Llama-13B} & 0.195 & \second{0.199} & 0.185 & \first{0.2}\\
\textbf{Llama-30B} & 0.287 & \first{0.294} & \second{0.289} & 0.283\\
\\ 
\bottomrule
\end{tabular}
}

\caption{Comparison with Coder-Reviewer Reranker.  Best results are colored in \first{first}, \second{second}.}
\label{table:crr}
\end{table*}

\subsection{Improvements are consistent across different generation temperatures}

In Figure~\ref{fig:temp_mbpp} (Supplement) we show how UCS reranking behaves for MBPP as the decoding sampling temperature increases. While accuracy can vary across temperatures, the ranking of the different methods remains consistent. Consensus-WUCS dominates in terms of accuracy for most of the temperature regimes until you hit the temperature of 1. Importantly, for lower temperatures where we get the best results, Both Consensus-WUCS as well as WUCS get the best accuracy. While just UCS is on par with mean log-probability ranking until a temperature of 0.4 after which it falls behind, we note that UCS does not use any probability information about the generation and thus a fair comparison would be to that of random ranking which it is consistency better than for almost the entire temperature range.

\subsection{Varying the maximum n-gram length does not change results\label{ngram_varying}}

\begin{figure}[h]
\captionsetup{justification=centering}
  \centering
  \begin{minipage}[b]{0.5\textwidth}
  \includegraphics[width=\textwidth]{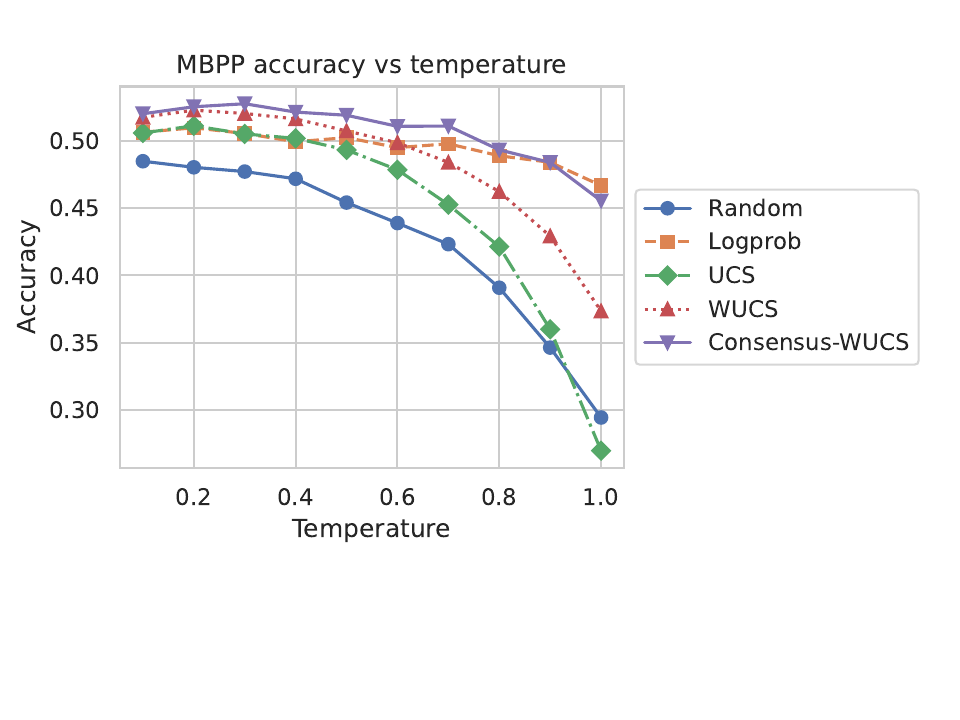}
  \vspace{-0.7in}
  \caption{\small{Accuracy for MBPP as the decoding sampling temperature increases.}}
  \label{fig:temp_mbpp}
  \end{minipage}
  \hspace{-0.2in}
  \hfill
  \begin{minipage}[b]{0.5\textwidth}
  \includegraphics[width=\textwidth]{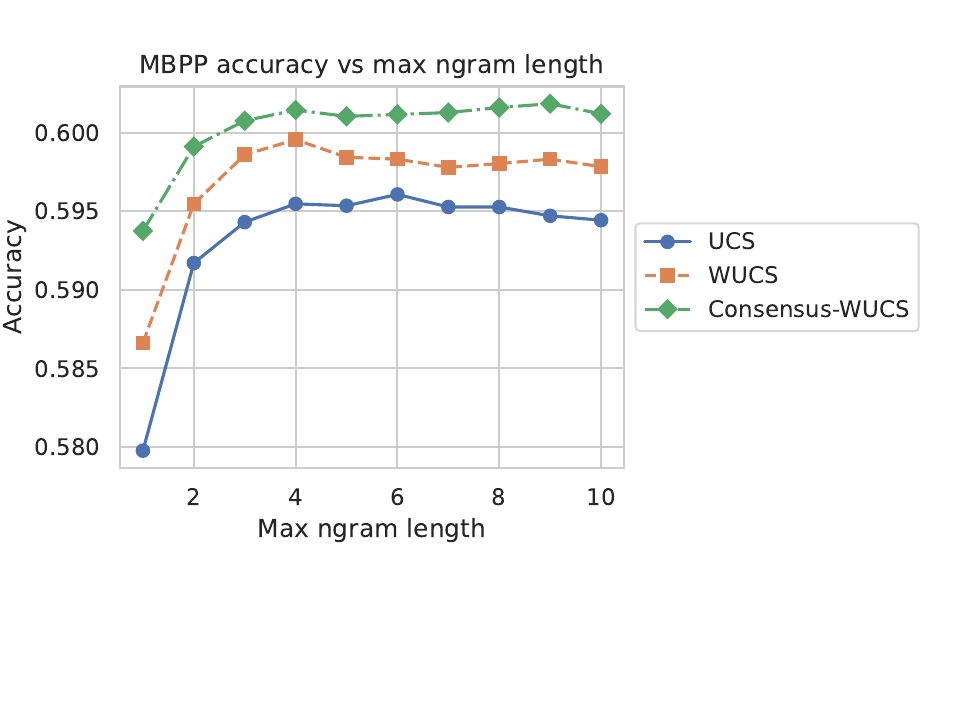}
  \vspace{-0.7in}
  \caption{\small{Accuracy for MBPP as the n in n-gram increases.}}
  \label{fig:ngram_mbpp}
  \end{minipage}
\end{figure}

As mentioned in Section~\ref{method}, UCS only considers unigrams. Here we consider Ngram Consistency Score -- the more generalized version. To account for the fact that a sentence will have fewer n-grams, the more $n$ increases, we multiply $p(t_j^{i,k})$ by $\frac{|g_i|}{|g_i|-|t_j^{i,k}|-1}$ where $t_j^{i,k}$ is now the $k$th appearance of the $j$th n-gram in the $i$th generation. In Figure~\ref{fig:ngram_mbpp} (Supplement), we show how the ranking behaves as the $n$ increases. As can be seen, while there is a slight improvement going from $n=1$ to $n=4$, the improvement flattens after that point. 4-grams is also what is conventionally used when computing BLEU score so it is interesting that the same value ends up being optimal in the drastically different setting of code generation with each word being a token instead of an English word.

\subsection{Increasing number of samples maintains reranking strength}

\begin{figure}
\captionsetup{justification=centering}
  \centering
  \begin{minipage}[b]{0.5\textwidth}
  \includegraphics[width=\textwidth]{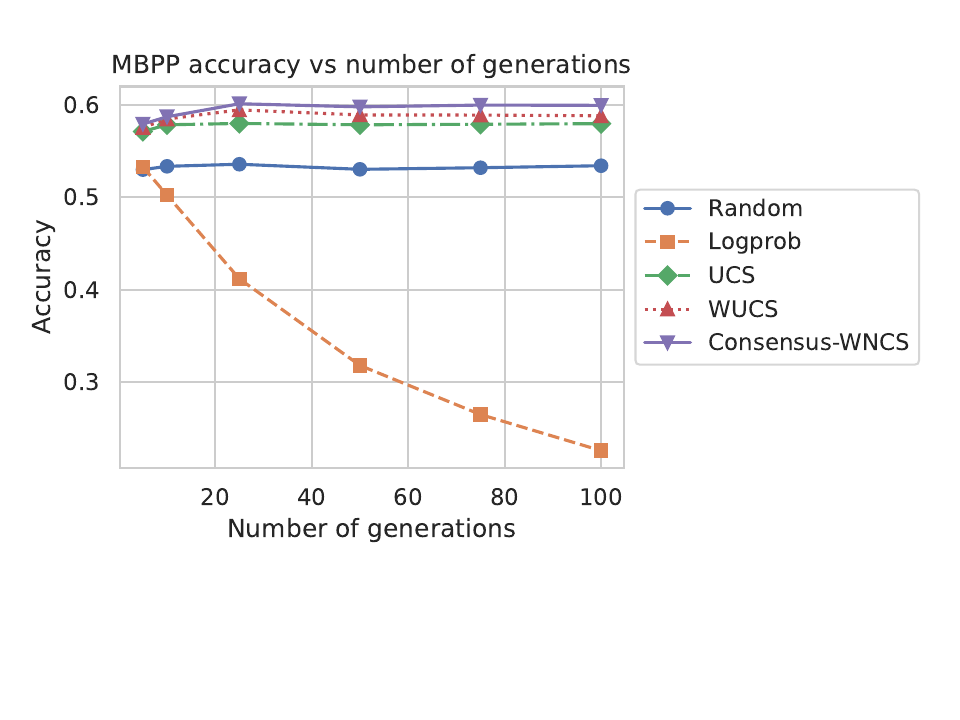}
  \vspace{-0.7in}
  \end{minipage}
  \hspace{-0.2in}
  \hfill
  \begin{minipage}[b]{0.5\textwidth}
  \includegraphics[width=\textwidth]{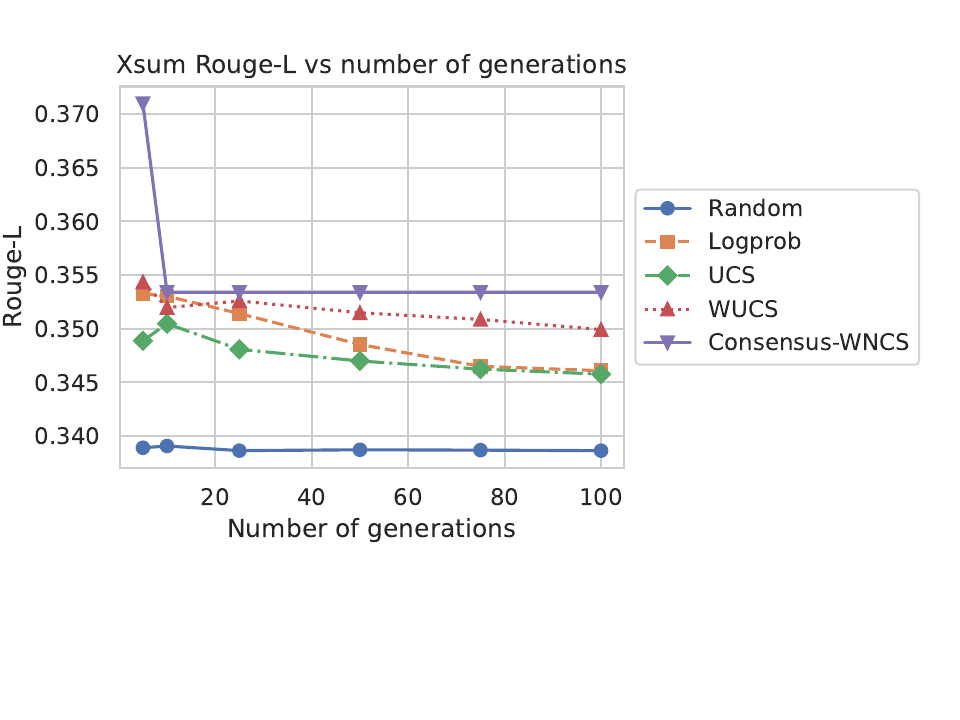}
  \vspace{-0.7in}
  \end{minipage}
  \caption{\small{Accuracy for MBPP and Rouge-2 for Xsum as the number of generations increase.}}
  \label{fig:sample_size}
\end{figure}

In Figure~\ref{fig:sample_size} (Supplement), we show how the performance changes for MBPP and Xsum as the number of samples increases. All variants of UCS are able to maintain accuracy (although Consensus-WUCS sees a drop in the beginning for Xsum but maintains its performance subsequently) even as the number of samples increases from 5 to 100. Meanwhile, the mean log probability ranking drastically declines in terms of accuracy, quickly falling below even random selection. This is likely due to the tendency of mean log probability ranking to choose degenerate sequences~\cite{holtzman2019curious} which UCS variants seem to be able to avoid.

\subsection{Ada model embeddings also give a boost\label{ada_embed_analysis}}

To understand how generalizable the intuition behind the GCS metric (as opposed to the UCS metric) is for other similarity functions, we took the generations and used the text-ada-embedding-002 model by OpenAI to generate embedding vectors for the generations. We then used cosine similarity between the generations as the similarity function and used $GCS_{\text{Cosine Similarity}}$ to rank. The results are in Table~\ref{table:ada}. Using OpenAI embeddings as well results in improved performance over Random selection as well as mean log probability ranking validating our intuition that choosing the generation that is on average, the most similar to all other generations is a good ranking metric. That said, this particular similarity function underperforms UCS, especially for code generation so we did not investigate it further.

\begin{table*}
\centering
\fontsize{8.25pt}{8.25pt}\selectfont
\setlength\tabcolsep{6.25pt} 
\scalebox{0.9}{
\begin{tabular}{p{4cm}ccccc}
\toprule
 & \textbf{Random} & \textbf{Mean-logp} & $\mathbf{GSC_{Ada}}$ & \textbf{Consensus-WUCS}\\
\textbf{HumanEval} & 0.437 & \second{0.533} & 0.487 & \first{0.568}\\
\textbf{MBPP} & 0.533 & 0.416 & \second{0.579} & \first{0.594}\\
\textbf{MBBP-S} & 0.549 & 0.568 & \first{0.601} & \second{0.589}\\
\textbf{MiniF2F (BLEU)} & 0.558 & 0.556 & \first{0.584} & \second{0.562}\\
\textbf{Xsum (Rouge-2)} & 0.197 & 0.214 & \first{0.219} & \first{0.219}
\\ 
\bottomrule
\end{tabular}
}
    \caption{Performance of cosine similarity of ada embedding as the similarity function. Metric is accuracy for HumanEval, MBPP, MBPP-S and BLEU for MiniF2F. Best results are colored in \first{first}, \second{second}.}
    \label{table:ada}
\end{table*}

\subsection{Normalizing inner product degrades performance\label{normalization_analysis}}

Neural generation models are well known to generate repetitive sequences~\cite{zhang2022coder, welleck2019neural}. In~\cite{welleck2019neural}, they modify the standard log-likelihood object for language models to minimize the probability of tokens immediately preceding the current token. This effectively pushes the model to generate unique new tokens and they show significant improvements in their model after they do this. If we normalize the inner product, then we would be effectively "canceling out" the contribution to the similarity score by having more unique tokens.

We evaluated the effect of normalizing the inner product by the vector norms. To understand better whether our performance is just an effect of selecting longer and more diverse sequences or whether the similarity metric itself is useful as well, we ran ablations where we evaluated ranking based on the longest sequence, as well as based on mean across the elements of $\mathbf{v}_i$ as defined in Section~\ref{method} -- which takes into account the sequence diversity. The results are in Table~\ref{table:normalization} in the Supplement. Normalization results in a decline in performance.  Furthermore neither ranking by the longest sequence nor ranking by sequence diversity is sufficient to give the results we see as neither result in a consistent improvement even against the Random selection baseline.

\begin{table}
\fontsize{8.25pt}{8.25pt}\selectfont
\setlength\tabcolsep{6.25pt} 
\scalebox{0.9}{
\begin{tabular}{p{4cm}cccccc}
\toprule
 & \textbf{Random} & \textbf{WUCS} & \textbf{WUCS-normalized} & \textbf{Longest} & \textbf{Most Diverse}\\
\midrule
 & \textbf{HumanEval}\\
\textbf{Codex002} & 0.435 & \first{0.558} & 0.462 & 0.441 & \second{0.51}\\
\textbf{Codex001} & 0.345 & \first{0.426} & \second{0.382} & 0.338 & 0.369\\
\textbf{Llama-30B} & 0.207 & \first{0.263} & \second{0.235} & 0.208 & 0.215\\
\midrule
 & \textbf{Random} & \textbf{WUCS} & \textbf{WUCS-normalized} & \textbf{Longest} & \textbf{Most Diverse}\\
\midrule
 & \textbf{MBPP}\\
\textbf{Codex002} & 0.536 & \first{0.587} & \second{0.576} & 0.529 & 0.52\\
\textbf{Codex001} & 0.475 & \first{0.52} & \second{0.517} & 0.475 & 0.457\\
\textbf{Llama-30B} & 0.262 & \first{0.287} & \second{0.278} & 0.263 & 0.245\\
\midrule
 & \textbf{Random} & \textbf{WUCS} & \textbf{WUCS-normalized} & \textbf{Longest} & \textbf{Most Diverse}\\
\midrule
 & \textbf{Xsum}\\
\textbf{Codex002} & 0.197 & \first{0.215} & \second{0.211} & 0.197 & 0.188\\
\textbf{Llama-30B} & 0.107 & \first{0.122} & \second{0.12} & 0.107 & 0.116\\
\textbf{GPT-J} & 0.065 & \first{0.07} & \first{0.07} & 0.065 & 0.069
\\ 
\bottomrule
\end{tabular}
}

    \caption{Impact of normalization. Best results are colored in \first{first}, \second{second}.}
    \label{table:normalization}
\end{table}

\begin{table*}
\fontsize{8.25pt}{8.25pt}\selectfont
\setlength\tabcolsep{6.25pt} 
\scalebox{0.9}{
\begin{tabular}{p{2cm}ccccccc}
\toprule
 & \textbf{HumanEval} & \textbf{MBPP} & \textbf{Xsum} & \textbf{MiniF2F} & \textbf{WMT-14 French To English} & \textbf{WMT14 German to English} \\
 Ratio & 1.95 & 1.34 & 1.21 & 1.08 & 1.07 & 1.08 \\
 \bottomrule
\end{tabular}
}
\caption{\small{Diversity ratio between best and worst generations from Codex002 model for various datasets}} \label{table:diversity}
\end{table*}

\begin{table}[htb]
\hspace{0.2in}
\begin{minipage}{0.43\textwidth}
\centering
\fontsize{8.25pt}{8.25pt}\selectfont
\setlength\tabcolsep{6.25pt} 
\scalebox{0.9}{
\begin{tabular}{p{2cm}cccccc}
\toprule
{\scriptsize \textbf{logprobs used}} \\
\toprule
& \textbf{Medoid} & \textbf{Mean-logp} & \textbf{WUCS} & \textbf{Consensus-WUCS}\\
\midrule
 & \textbf{HumanEval}\\
\textbf{Codex002} & 0.437 & 0.539 & \second{0.558} & \first{\textit{0.568}}\\
\textbf{Codex001} & 0.354 & 0.408 & \second{0.426} & \first{\textit{0.445}}\\
\textbf{Code-Cushman} & 0.335 & 0.355 & \second{0.373} & \first{\textit{0.381}}\\
\textbf{Llama-13B} & 0.17 & 0.17 & \second{0.187} & \first{\textit{0.192}}\\
\textbf{Llama-30B} & 0.225 & 0.228 & \second{0.263} & \first{\textit{0.267}}\\
\midrule
 & \textbf{MBPP-S}\\
\textbf{Codex002} & \second{0.583} & 0.57 & 0.580 & \first{\textit{0.589}}\\
\textbf{Codex001} & 0.532 & 0.515 & \second{0.535} & \first{\textit{0.546}}\\
\textbf{Code-Cushman} & 0.467 & 0.456 & \second{0.472} & \first{\textit{0.488}}\\
\textbf{Llama-13B} & \first{0.284} & 0.27 & 0.266 & \second{\textit{0.277}}\\
\textbf{Llama-30B} & 0.357 & 0.348 & \second{0.363} & \first{\textit{0.373}}\\
\midrule
 & \textbf{MBPP}\\
\textbf{Codex002} & 0.563 & 0.512 & \second{0.587} & \first{\textit{0.594}}\\
\textbf{Codex001} & 0.505 & 0.503 & \second{0.520} & \first{\textit{0.525}}\\
\textbf{Code-Cushman} & 0.343 & 0.319 & \second{0.405} & \first{\textit{0.420}}\\
\textbf{Llama-13B} & \first{\textit{0.202}} & 0.197 & 0.195 & \second{0.199}\\
\textbf{Llama-30B} & 0.276 & 0.273 & \second{0.287} & \first{\textit{0.294}}\\
\\ 
\bottomrule
\end{tabular}
}
\end{minipage}

\captionsetup{justification=centering}
\caption{\small{Accuracy of generated code for HumanEval, MBPP, MBBP-S}. Best results are colored in \first{first}, \second{second}. Italics for best in category (logprobs used vs not)}\label{table:code_gen_accuracy}
\end{table}

\begin{table*}[htb]
\hspace{0.2in}
\begin{minipage}{0.5\textwidth}
\centering
\fontsize{8.25pt}{8.25pt}\selectfont
\setlength\tabcolsep{6.25pt} 
\scalebox{0.9}{
\begin{tabular}{p{2cm}ccccc}
\toprule
{\scriptsize \textbf{logprobs used}} \\
\toprule
 & \textbf{Medoid} & \textbf{Mean-logp} & \textbf{WUCS} & \textbf{Consensus-WUCS}\\
\midrule
 & \textbf{MiniF2F BLEU}\\
\textbf{Codex002} & \first{58.2} & 52.9 & 55.8 & \second{56.2} \\
\textbf{Llama-13B} & \first{24.9} & 24.2 & 24.7 & \second{24.8}\\
\textbf{Llama-30B} & \first{26.4} & 25.6 & 25.7 & 25.7\\
\textbf{GPT-J} & \first{24.8} & 24 & \first{24.8} & \first{24.8}\\
\midrule
 & \textbf{Xsum Rouge2}\\
\textbf{Codex002} & \second{21.8} & 21.4 & 21.5 & \first{21.9}\\
\textbf{Llama-13B} & 10.3 & 10.3 & \first{10.6} & \first{10.6}\\
\textbf{Llama-30B} & 12 & 12.2 & \second{12.2} & \first{12.3}\\
\textbf{GPT-J} & \second{6.9} & 6.6 & \first{7} & \second{6.9}\\
\midrule
 & \textbf{Xsum RougeL} \\
\textbf{Codex002} & \first{36.3} & 35.1 & 35.3 & \second{35.6}\\
\textbf{Llama-13B} & 20.7 & 20.3 & \first{21} & \second{20.9}\\
\textbf{Llama-30B} & 22.7 & 22.8 & \first{23.1} & \first{23.1}\\
\textbf{GPT-J} & \second{17.5} & 16.6 & \first{17.8} & \second{17.5} \\
\midrule
& \textbf{WMT14 French $\rightarrow$} \\ & \textbf{English BLEU} \\
\textbf{Codex002} & 35.9 & \second{36.6} & 36.5 & \first{37} \\
\textbf{Llama-13B} & 4.2 & \second{4.5} & \second{4.5} & \first{4.6} \\
\textbf{Llama-30B} & 4 & 4 & \first{4.1} & \first{4.1} \\
\textbf{GPT-J} & 3.8 & 3.9 & \first{4} & \first{4} \\
\midrule
& \textbf{WMT14 German $\rightarrow$} \\ & \textbf{English BLEU} \\
\textbf{Codex002} & 31.2 & \second{33.2} & 32.1 & \first{34} \\
\textbf{Llama-13B} & 3.1 & \first{4} & 3.5 & \second{3.6} \\
\textbf{Llama-30B} & 3.5 & \first{3.9} & 3.8 & \first{3.9} \\
\textbf{GPT-J} & 3.2 & 3.2 & \first{3.3} & \first{3.3} \\
\bottomrule
\end{tabular}
}
\end{minipage}

\end{table*}

\begin{table}
\captionsetup{justification=centering}
\centering
\begin{tabular}{|lccc|}
\hline
 & \textbf{Aqua} & \textbf{Multiarith} & \textbf{StrategyQA}\\
\midrule
\textbf{Codex001} & +2.8\% & +4.4\% & +2.5\% \\
\textbf{Codex002} & - & +7.1\% & +3.3\% \\
\textbf{LaMDA-137} & +1.9\% & +4.4\% & +3.9\% \\
\textbf{UL2-20B} & -1\% & -0.1\% & -0.1\% \\
\hline
\end{tabular}
\caption{\small{Ratio of average GSC score for correct generations by average GSC score for incorrect generations.}}\label{table:UCS_correct_incorrect}
\end{table}

\end{document}